\documentclass{ieeeaccess}
\usepackage{cite}
\usepackage{amsmath,amssymb,amsfonts}
\usepackage{algorithmic}
\usepackage{graphicx}
\usepackage{textcomp}
\usepackage[hidelinks]{hyperref}
\usepackage{subcaption}
\usepackage{xcolor}
\usepackage{bm}
\DeclareMathOperator*{\argmin}{arg\,min}
\usepackage[switch]{lineno}
\usepackage{amsthm}
\newtheorem*{remark}{Remark}

\makeatletter
\AtBeginDocument{\DeclareMathVersion{bold}
\SetSymbolFont{operators}{bold}{T1}{times}{b}{n}
\SetSymbolFont{NewLetters}{bold}{T1}{times}{b}{it}
\SetMathAlphabet{\mathrm}{bold}{T1}{times}{b}{n}
\SetMathAlphabet{\mathit}{bold}{T1}{times}{b}{it}
\SetMathAlphabet{\mathbf}{bold}{T1}{times}{b}{n}
\SetMathAlphabet{\mathtt}{bold}{OT1}{pcr}{b}{n}
\SetSymbolFont{symbols}{bold}{OMS}{cmsy}{b}{n}
\renewcommand\boldmath{\@nomath\boldmath\mathversion{bold}}}
\makeatother

\def\BibTeX{{\rm B\kern-.05em{\sc i\kern-.025em b}\kern-.08em
    T\kern-.1667em\lower.7ex\hbox{E}\kern-.125emX}}

\newcommand{\revision}[1]{{\color{black} #1}}

\renewcommand{\thevol}{00}
\renewcommand{\theyear}{0000}

\begin{document}
\history{Date of publication xxxx 00, 0000, date of current version xxxx 00, 0000.}
\doi{n/a}

\title{A Unification Between Deep-Learning Vision, Compartmental Dynamical Thermodynamics, and Robotic Manipulation for a Circular Economy}
\author{\uppercase{Federico Zocco}\authorrefmark{1},
\uppercase{Wassim M. Haddad}\authorrefmark{2}, \uppercase{Andrea Corti}\authorrefmark{3}, and \uppercase{Monica Malvezzi}\authorrefmark{3}}

\address[1]{Centre for Sustainable Manufacturing and Recycling Technologies (SMART), Wolfson School of Mechanical, Electrical and Manufacturing Engineering, Loughborough University, England, United Kingdom (e-mail: federico.zocco.fz@gmail.com)}
\address[2]{School of Aerospace Engineering, Georgia Institute of Technology, Atlanta, GA, USA (e-mail: wassim.haddad@aerospace.gatech.edu)}
\address[3]{Department of Information Engineering and Mathematics, University of Siena, Italy (e-mail: \{andrea.corti, monica.malvezzi\}@unisi.it}
\tfootnote{This work was not supported by any organization.}

\markboth
{F. Zocco \headeretal: A Unification Between Research Topics for a Circular Economy}
{F. Zocco \headeretal: A Unification Between Research Topics for a Circular Economy}

\corresp{Corresponding author: Monica Malvezzi (e-mail: monica.malvezzi@unisi.it)}

\begin{abstract}
The shift from a linear to a circular economy has the potential to simultaneously reduce uncertainties of material supplies and waste generation. \revision{However, to date, the development of robotic and, more generally, autonomous systems have been rarely integrated into circular economy implementation strategies despite their potential to reduce the operational costs and the contamination risks from handling waste. In addition, the science of circularity still lacks the physical foundations needed to improve the accuracy and the repeatability of the models. Hence, in this paper, we merge deep-learning vision, compartmental dynamical thermodynamics, and robotic manipulation into a theoretically-coherent physics-based research framework to lay the foundations of circular flow designs of materials. The proposed framework tackles circularity by generalizing the design approach of the Rankine cycle enhanced with dynamical systems theory. This differs from state-of-the-art approaches to circular economy, which are mainly based on data analysis, e.g., material flow analysis (MFA). We begin by reviewing the literature of the three abovementioned research areas, then we introduce the proposed unified framework and we report the initial application of the framework to plastics systems along with initial simulation results of reinforcement-learning control of robotic waste sorting. This shows the framework applicability, generality, scalability, and the similarity and difference between the optimization of artificial neural systems and the proposed compartmental networks. Finally, we discuss the still not fully exploited opportunities for robotics in circular economy and the future challenges in the theory and practice of the proposed circularity framework.}   
\end{abstract}


\titlepgskip=-21pt

\maketitle

\section{Introduction}
\label{sec:intro}
\PARstart{V}{isualize} for a moment that a built environment and human activities are contained within a unit or compartment. This compartment has an input flow and an output flow of materials: the former comes from natural reserves, whereas the latter is the generation of waste and pollution. The central issue with the input flow of the compartment is that the supplies of several raw materials have a high level of risk and are also needed to manufacture high-tech and green technologies essential to a modern society \cite{CRMall}. Hence, these materials are labeled as ``critical'' by the European Union \cite{CRM-EU} and the United States \cite{CRM-US}. 

The central issue with the output flow of the compartment is that environmental pollution is gradually altering the balance of ecosystems. Climate change is a manifestation of this alteration and it is mainly caused by the accumulation of greenhouse gases in the atmosphere such as carbon dioxide \cite{NASA-CO2}. Other manifestations are the rise of the global temperature and the sea levels, and more frequent extreme weather events and patterns such as hurricanes or heat waves (depending on the region) \cite{NASA-effects}. In a 2024 report, the United Nations projected 2.684, 3.229, and 3.782 billion tons of waste generated in 2030, 2040, and 2050, respectively, if urgent action is not taken \cite{UN2024}. The Organization for Economic Co-operation and Development  (OECD) reported that the share of municipal solid waste landfilled in the OECD area has decreased from an estimated 53\% to 40\% between 2000 and 2021, with some countries no longer using landfills (Switzerland, Germany, Denmark, Finland, Sweden, and Japan) \cite{OECD2023}.

\revision{In this context, the circular economy (CE) paradigm aims in minimizing both the input and the output flows of the unit or compartment through an increase of maintenance, reuse, remanufacturing, recycling, and material efficiency \cite{EllenMAF,russell2023value}. Keeping materials in use within the economy (i.e., the unit) for a longer time reduces the extraction of finite resources and it also could increase the self-sufficiency of the communities \cite{mederake2023without}. According to a 2021 report from the Centre for European Policy Studies (CEPS), there are several barriers to the adoption of circularity practices such as existing policies, economic factors, supply chains, technology, and consumer preferences \cite{CEPSonCE}. From a scientific and educational perspective, the principles of circularity are very rarely integrated into highly technical programs and courses, which further delays their adoption by the general public and younger generations \cite{kirchherr2019towards,mesa2021towards,giannoccaro2021features}. 

One of the main reasons for the weak intersection between technical subjects and circularity is the lack of solid physical and engineering foundations of the latter since, traditionally, CE is a topic of business and environmental sciences. Hence, there is a societal need for integrating CE principles into technically rigorous programs while still maintaining their high mathematical and physics standards. Indeed, such standards and skills are essential to create accurate circularity models and to design and prototype circularity-oriented devices and intelligent systems. 

In this context, this paper makes the following contributions:}
\begin{itemize}
\item{We merge deep-learning vision, compartmental dynamical thermodynamics, and robotic manipulation into a theoretically-coherent research framework. Our framework has a double benefit: namely, it integrates robotics into the systemic perspective needed to address circular economic challenges; and it improves the \revision{physical foundations} of the theory of CE.}
\item{\revision{We report the initial application of the proposed framework to plastics systems (Section \ref{sec:ForPlastics}).}}
\item{We discuss opportunities for robotics in a circular economy to address the existing gap between Industry 4.0 techniques and Sustainability \cite{bakshi2022sustainability} \revision{and, from an educational point of view, to address the gap between robotics and environmental programs (Section \ref{sec:oppRobInCE})}.}
\end{itemize}

The rest of the paper is organized as follows. Section \ref{sec:4topics} covers the literature review relevant for our theoretically-coherent research framework by focusing on four different topics. Then, in Section \ref{sec:combine}, we present a unification of the four key topics discussed in Section \ref{sec:4topics}, whereas in Section \ref{sec:library} we collect existing models from the literature aligned with our framework. In Section \ref{sec:oppRobInCE}, we focus on the opportunities for robotics in a circular economy, \revision{while in Section \ref{sec:ForPlastics} we apply the proposed framework to plastics systems}. Finally, in Section \ref{sec:concl}, we give conclusions.

\section{THREE REPRESENTATIVE TOPICS FOR CIRCULAR ECONOMY}\label{sec:4topics}
This section covers the literature regarding the four topics mentioned in the title and intersected in this paper; namely, circular economy (CE), compartmental dynamical thermodynamics (CDT), robotic manipulation (RM), and deep-learning vision (DLV).

\subsection{Circular Economy}\label{sub:CE}
Velenturf \textit{et al.} \cite{velenturf2021principles} view circular economy as ``[...] an emerging practical ideology that lacks an evidence-based theoretical framework to guide implementation [...]''. Kirchherr \textit{et al.} \cite{kirchherr2017conceptualizing} identified 114 definitions of circular economy to highlight the lack of clarity and agreement about its meaning. This lack of evidence-based and clear foundations is one of the main motivations for this paper. Here, we adopt the popular and intuitive definition of the Ellen MacArthur Foundation \cite{EllenMAF}, which looks at closing the loops of material flows similarly to what we discussed in Section \ref{sec:intro}. The large number of definitions of CE goes in parallel with the large number of indicators developed to measure circularity \cite{moraga2019circular,saidani2019taxonomy,zocco2022circularity}. For example, De Pascale \textit{et al.} \cite{de2021systematic} identified 61 indicators.

The interplay between CE and Industry 4.0 is another frequent topic in the literature. Overall, the gap between the two realms is still existing and the latter has the potential to accelerate the actualization of the former via, for example, big data, machine learning, internet-of-things, robotics, and automation \cite{rosa2020assessing}. In particular, Laskurain-Iturbe \textit{et al.} \cite{laskurain2021exploring} identified additive manufacturing and robotics as the most promising technologies. Kumar \textit{et al.} \cite{kumar2024barriers} identified ``high investment'' and ``technological immaturity'' as two of the factors hindering the adoption of Industry 4.0 to enhance circularity in the food sector. Trust and common ambitions between members of CE projects were identified as very important in \cite{leising2018circular} to establish the required new collaborations between supply chain players, while the adoption of CE for small-medium enterprises was studied by Howard \textit{et al.} \cite{howard2022going}. 

\revision{To date, material flow analysis (MFA) is the main methodology for assessing circularity \cite{brunner2016handbook,CSIRO-MFA,EU-MFA,cullen2022material}. It is essentially based on the analysis of real data of material stocks and flows recorded over time. In terms of its physical foundations, MFA is based on the mass balance and the MFA results are usually represented with Sankey diagrams with arrows indicating the direction of the flows between different life-cycle stages: e.g., extraction and manufacturing \cite{brunner2016handbook,CSIRO-MFA,EU-MFA,cullen2022material}. With respect to MFA, the framework proposed in this paper, firstly, adds thermodynamic laws to the mass balances and applies them to the control surfaces used in the design of thermodynamic cycles; and secondly, it makes use of dynamical system models (i.e, ordinary differential and difference equations) instead of data modeling and analysis.}

\subsection{Compartmental Dynamical Thermodynamics}\label{sub:CDT}
Thermodynamics is universal, and hence, in principle, it applies to everything in nature---from simple engineering systems to complex living organisms to our expanding universe. The laws of thermodynamics form the theoretical underpinning of virtually all disciplines in science and engineering, and hence, it is not surprising that thermodynamics can be used to model, plan, and control circular material flows \cite{zocco2023thermodynamical}. The fundamental and unifying concept in the analysis of a circular economy is the concept of energy flow and material flow. The energy of a state of a dynamical system is the measure of its ability to produce changes (motion) in its own system state as well as changes in the system states of its surroundings. These changes occur as a direct consequence of the energy flow between different subsystems (or compartments) within the dynamical system.

In \cite{haddad2009thermodynamics,haddad2013unification,haddad2017thermodynamics,haddad2020condensed,haddad2019dynamical}, the author merges classical thermodynamics with dynamical systems theory to develop a post-modern framework for thermodynamics predicated on open (i.e., input-state-output) interconnected dynamical systems that exchange matter and energy with the environment in accordance with the first law (conservation of energy) and the second law (non-conservation of entropy) of thermodynamics. Using a dynamical systems perspective, \cite{haddad2019dynamical} provides a system-theoretic framework for non-equilibrium thermodynamics using a state-space formalism involving nonlinear compartmental dynamical system models characterized by energy conservation laws that are consistent with basic thermodynamic principles. This is in sharp contrast to classical thermodynamics wherein a static input-output description of the system is used. If a physical system possesses conservation properties externally, then there exists a high probability that the system also possesses these same properties internally. Specifically, if the system possesses conservation properties internally and does not violate any input-output behavior of the physical system established by experimental evidence, then the state space model is theoretically credible.

A state-space formulation is thus essential for developing a thermodynamic model with enough detail for describing the thermal behavior of heat and dynamical energy. In addition, such a model is crucial in accounting for internal system properties captured by compartmental system dynamics characterizing conservation laws, wherein subsystem energies and matter can only be transported, stored, or dissipated but not created. Furthermore, an input-state-output model is ideal in capturing system interconnections with the environment involving the exchange of matter, energy, or material, with inputs serving to capture the influence of the environment on the system, outputs serving to capture the influence of the system on the environment, and internal feedback interconnections---via specific output-to-input assignments---serving to capture interactions between subsystems. The state space model additionally enforces the fundamental property of causality, that is, nonanticipativity of the system, wherein future values of the system inputs do not influence past and present values of the system outputs. Thus, compartmental dynamical thermodynamics is ideal in providing a dynamical systems framework for modelling, planning, and controlling circular material flows in a circular economy.

\subsection{Robotic Manipulation for Reprocessing}\label{sub:RM}
Processing waste items in a circular economy perspective with automated and robotic systems presents several challenges. Kiyokawa \textit{et al.}  \cite{kiyokawa2022challenges} identified three main elements that are needed in sorting automated systems: $i$) the end effector, which has to grasp and manipulate objects with different shapes and properties, often in a dirt and contaminated environment; $ii$) a sensing system, often based on vision, able to recognize the shape, position, category, and property of the waste item to be processed; and $iii$) a system able to plan and control stable and efficient trajectories and operation sequences.

Grasping and manipulating objects is one of the most relevant and challenging abilities required for robotic systems in several applications \cite{billard2019trends}. While traditional grasp models were based on simplified assumptions, often considering rigid objects with a limited number of contacts, approximated as points in \cite{prattichizzo2016grasping}, current robotic gripper developments and applications exploit more complex conditions. A rigid approach both in modeling and in the design of robotic grippers and end effectors is not suitable for getting simple, robust, effective, and reliable solutions in unstructured environments. The current trend in robotic end effector developments is towards non-rigid, or \textit{soft} structures, as reviewed by Zaidi \textit{et al.} \cite{zaidi2021actuation}. Such systems are able to perform complex tasks, for example grasping and manipulating deformable objects, as the device presented by Marullo \textit{et al.} \cite{marullo2020mag}, or multiple objects, as described by Li \textit{et al.} \cite{li2024grasp}. Tactile sensing is important to detect grasp conditions and stability, especially in inherently underactuated soft grippers \cite{qu2023recent}. Soft robotic grippers can implement novel and interesting grasping strategies, for example exploiting environmental constraints, as introduced by Eppner \textit{et al.} \cite{eppner2015exploitation} and implemented by Pozzi \textit{et al.} \cite{pozzi2020hand}. Physical constraints guiding grasping tasks can be embedded as specific features in soft robotic grippers and hands \cite{bo2022automated}.
All these challenges (unstructured workspace, presence of dirt and contaminants, and deformable and fragile objects) are greatly present in systems for waste manipulation \cite{ku2021deep}.

Notwithstanding  the increasing need for robotic systems in waste sorting, available  
robotic grippers and end effectors are often not designed to cope with the large variation in material types and shapes encountered in waste material sorting, often operating with shredded material streams. As underlined by Billard and Kragic in \cite{billard2019trends}, vacuum systems are the most efficient solutions when the task consists of grasping and removing an object from a set, without manipulating it; for example, in waste sorting applications. These systems are efficient and can easily adapt to different objects \cite{papadakis2020use,9216746,koskinopoulou2021robotic,um2023fast}. In \cite{9551603} an algorithm for defining the optimal suction point in plastic waste sorting is defined. 
Another solution able to comply with waste item variability, employing air flow, is presented by Engelen \textit{et al.} \cite{engelen2023high}. 

When the task requires dexterity or manipulation ability, multi-fingered robotic hands, grippers, or other end effectors are needed. 
The development of soft robotic grippers and hands, exploiting soft materials and underactuation, has represented in the last decade an effective solution for obtaining the suitable adaptability and versatility requirements needed in unstructured environments, with simple and robust devices. This type of robotic system also represents an interesting solution in waste management systems \cite{rithvik2023soft}.
Compliant materials and system underactuation provide adaptability to environmental uncertainties and system robustness, but lead to a new set of challenges in robot planning and control.  Specific control strategies, inspired by human behavior and based on environmental constraint exploitation, have been developed in \cite{pozzi2020hand,eppner2015exploitation}.
Robotic grippers introduced in \cite{achilli2023multibody,9530533} employ parts designed for smart environmental constraint exploitation \cite{achilli2023underactuated}. 
One interesting, bio-inspired structure implemented in soft grippers is the Fin Ray\textsuperscript{\textregistered} effect \cite{crooks2016fin}. Almanzor \textit{et al.} \cite{almanzor2022autonomous} developed a soft gripper based on this principle for litter sorting.

Different structures can be combined to improve gripper capabilities. For example, Sadeghian \textit{et al.} \cite{sadeghian2022vision} presented a trimodal gripper including a vacuum suction mechanism, a nano-polyurethane adhesive mechanism, and a set of claws is developed. Another interesting solution is presented in \cite{9981625}, where Kirigami robotic grippers are developed to manipulate different objects, including items in waste sorting systems. Bonello \textit{et al.} \cite{bonello2017exploratory} introduced an automated system for dry waste sorting and for this purpose developed a versatile gripper for waste manipulation.
The design of a robotic gripper specifically developed for glass manipulation in waste sorting is presented by Kang \textit{et al.} in 
\cite{kang2024enhancing}.

Besides grasping and manipulating tasks, often in automated systems in the circular economy contexts involve more articulated operations.   One example is represented by disassembling elements \cite{daneshmand2023industry}, where complex bimanual operations are required; for example unscrewing, removing components, cutting, etc. \cite{tan2021hybrid,kay2022robotic,vongbunyong2015disassembly}. Disassembly operations are becoming particularly important in waste electronic and electrical equipment (WEEE) \cite{foo2022challenges}.
Autonomous systems have been proposed for disassembly tasks of specific items, e.g., electronic components \cite{farhan2021autonomous,marconi2019feasibility}, batteries \cite{hathaway2023towards}, cellphones \cite{figueiredo2018high}, and press fitted components \cite{huang2020case}.
For disassembling tasks, multifunctional grippers enabling bimanual tasks have been presented in \cite{schmitt2011disassembly,borras2018kit}.

Due to their complexity, completely automated systems are still not very diffused, while robotic systems are more frequently introduced in collaboration with human operators, as reviewed by Hjorth \textit{et al.} \cite{hjorth2022human}.
Examples of human/robot collaborative cells for disassembly tasks have been proposed in \cite{huang2021experimental,huang2020case}.
Both in autonomous and collaborative disassembly tasks, proper grasp planner algorithms \cite{fernandez2006part} and specific force sensing systems \cite{schumacher2014force} are defined.

\subsection{Deep-Learning Vision for Robotics}\label{sub:DLV} 
The breakthroughs in deep learning over the last decade have significantly improved the accuracy, speed, and generalization power of vision systems. More recently, the robotics community is leveraging such advances to enhance the ability of robots to navigate and interact with environments more complex than research testbed laboratory environments. Lee \textit{et al.} \cite{lee2020making} addressed contact-rich tasks by training a deep neural network on RGB, depth, force-torque, and proprioception data. \revision{The real robot is controlled by a reinforcement learning (RL) policy learned through self-supervision to perform peg insertion.} Vision-based navigation was developed in \cite{devo2020towards,bansal2020combining}; the former used a deep RL approach, whereas the latter combined model-based control with learning from images.

Object tracking is a useful ability for a robot that needs to interact with a dynamic environment. An object tracking system \revision{was developed by Huang \textit{et al.} \cite{huang2023tracking} using three novel modules, namely, the motion disentanglement network, the patch association network, and the patch memory network, while the system of Gordon \textit{et al.} \cite{gordon2018re} simultaneously tracks and updates a deep object tracker in real-time with a single forward pass. Other algorithms were proposed in \cite{fan2017parallel,kiani2017need,song2022transformer,richter2021robotic}}. A literature review on deep-learning-based visual tracking can be found in \cite{marvasti2021deep}. An effective technique for visual tracking is to use object detection on each video frame, but annotating data manually is very time consuming. Hence, Horv\'ath \textit{et al.} \cite{horvath2022object} automated the annotation process. Another important task for intelligent robots is estimating the pose of target objects, especially for their manipulation. Pose estimation of objects was developed in \cite{sundermeyer2018implicit,wang20206,li2018deepim,tan2023smoc}. 

Visual servoing typically consists of a combination of one or more cameras, the image processing module, and the control algorithms that translate the image information and other sensor data into robot actions. \revision{Al-Shanoon \textit{et al.} \cite{al2022robotic}, for example, trained a convolutional neural network using only synthetic data and they used a feedback control scheme based on the 3-D object pose for manipulation tasks.} Other approaches were proposed in \cite{hay2023noise,bateux2018training}, while Costante \textit{et al.} \cite{costante2020uncertainty} proposed a data-driven odometry module that provides information about the uncertainty of its estimations.             

Deep-learning vision has also been proposed for autonomous marine debris detection \cite{fulton2019robotic,zocco2023towards,hong2020trashcan}, whereas non-marine waste was the focus of \cite{bashkirova2022zerowaste,bai2018deep,wang2020vision,kiyokawa2021robotic}. \revision{In particular, Bashkirova \textit{et al.} \cite{bashkirova2022zerowaste} collected and published the first in-the-wild industrial-grade waste detection and segmentation dataset, namely, ZeroWaste, to facilitate the development of vision systems for highly cluttered recycling scenarios, while Kiyokawa \textit{et al.} \cite{kiyokawa2021robotic} opted for graspless push-and-drop and pick-and-release manipulation for their robotic waste sorter.}

Ku \textit{et al.} \cite{ku2021deep} used a deep neural network to find an effective grasping pose for sorting construction and demolition waste. Grasping waste items is particularly challenging since their shape is difficult to predict. In contrast, manufacturing processes have less uncertainties since the geometry of the products are known. A similar consideration holds for disassembly; the conditions of an end-of-life device are difficult to predict, which makes the automation of their disassembly sequence more challenging than assembling processes in manufacturing, where the assembly sequence is known. 

Jahanian \textit{et al.} \cite{jahanian2019see} focused on electronic waste for disassembly operations, whereas \cite{brogan2021deep,mangold2022vision,foo2021screw} detected the screws since their removal is the initial step of many disassembly sequences. Medical waste was considered in \cite{zocco2023visual,zocco2024towards}. Visual material recognition can improve both the effectiveness of robotic grasping (e.g., lower control gains with fragile items) and the automation of material sorting \revision{and mapping for a circular economy as extensively reviewed and discussed by Zocco \textit{et al.} \cite{zocco2022material} in proposing material measurement units, which are distributed autonomous visual recognizers of the materials located in a certain area of interest, e.g., a town}. Material recognition from images was performed \revision{by Bell \textit{et al.} \cite{bell2015material}. Specifically, they introduced a large-scale open dataset of materials in the wild, namely, MINC, and then, they trained convolutional neural networks to simultaneously perform material recognition and segmentation.} Other works on material recognition are \cite{schwartz2019recognizing,lagunas2019similarity}.  

Along with the type of material, the mass of an object is another useful property with a double application; the knowledge of the mass can make manipulation more adaptive (e.g., higher control gains with heavier items) and can also enable an autonomous quantification of material stocks to improve natural resources management in a circular economy context \cite{zocco2024synchronized,zocco2022material} (i.e., to automate the process of material quantification in a certain area). Mass estimation from images was proposed in \cite{standley2017image2mass,andrade2023improving,patel2022adding,diaz2022simultaneous}.

\section{UNIFICATION OF DEEP-LEARNING VISION, COMPARTMENTAL THERMODYNAMICS, ROBOTIC MANIPULATION, AND CIRCULAR ECONOMY}\label{sec:combine}
\begin{figure*}
    \centering
    \includegraphics[width=0.8\textwidth]{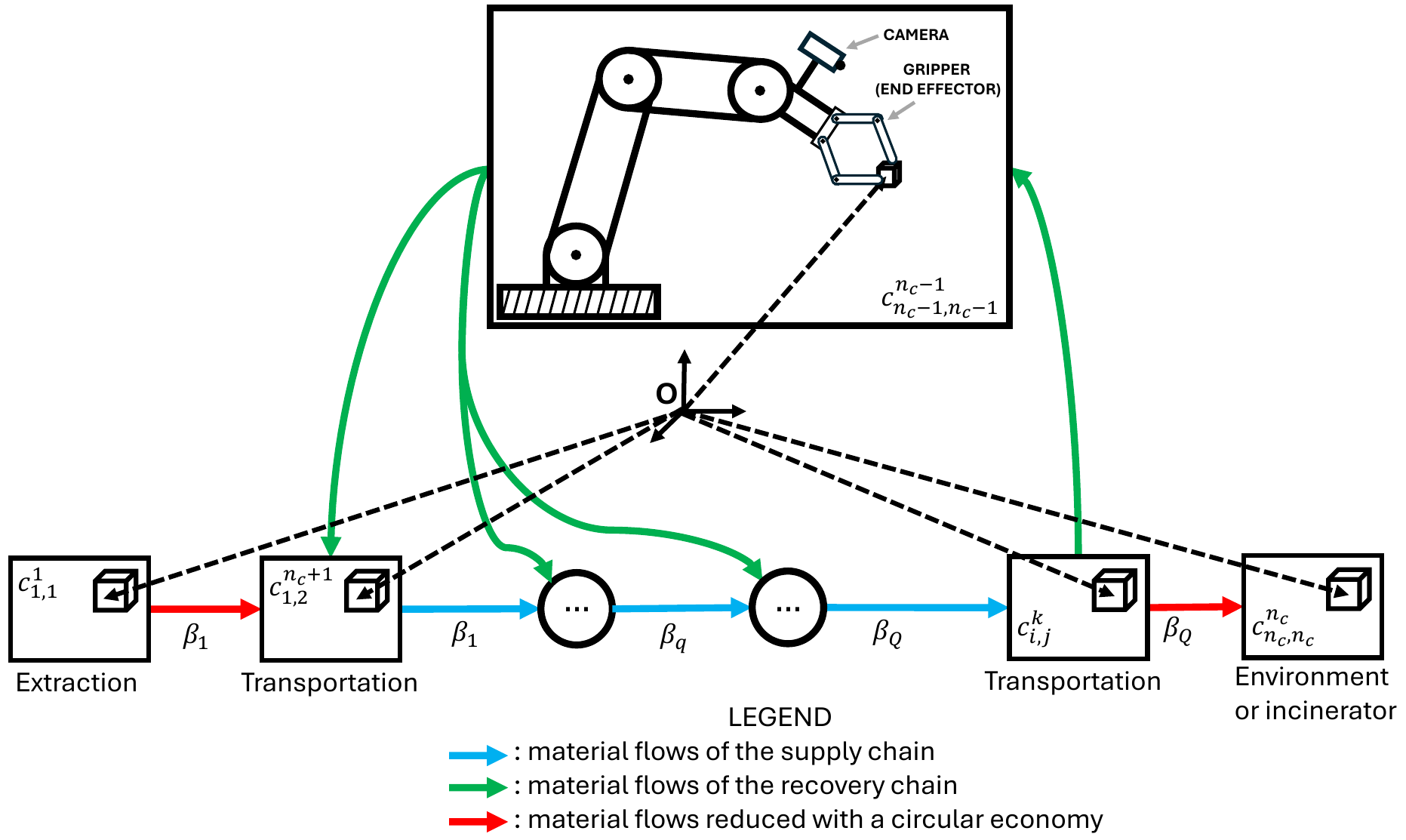}
    \caption{Systemic view for deep-learning vision, compartmental dynamical thermodynamics, and robotic manipulation coherently combined to design circular flows of materials. Nomenclature adopted from Zocco \textit{et al.} \cite{zocco2023thermodynamical}.}
    \label{fig:CombiningFigure}
\end{figure*}

In this section, we combine the four topics mentioned in the title and covered in Sections \ref{sub:CE}--\ref{sub:DLV}. The graphical intersection of the four topics is depicted in Fig. \ref{fig:CombiningFigure}, where each stage of the supply and recovery chain is a thermodynamic compartment indicated by $c^k_{i,j}$ following the nomenclature of \cite{zocco2023thermodynamical}. 
This representation of a life-cycle is referred to as ``compartmental diagram'' in \cite{zocco2023thermodynamical}. The origin $O$ of the frame of reference is the tail of the position vector of an infinitesimal cube of material which crosses the compartments. The compartments are connected by material flows (the arrows) and the material type is indicated by $\beta_q$, with $q \in [1, Q]$. The total number of compartments is $n_c$. The ``Transportation'' compartment usually does not modify the material, hence the input flow is of the same type as the output flow. The recovery chain can be automated to reduce the costs of waste sorting and re-manufacturing operations \cite{zocco2024towards}. The robot can be equipped with a camera and a gripper for handling waste materials or end-of-first-use products. Then, the reprocessed materials or products can reenter the supply chain at different stages; recycled material can be used by a manufacturer to reduce its dependence on finite natural resources, while a repaired product can be transported directly to a new consumer.           

\textbf{Topic 1 (CE):} The Rankine cycle is one of the simplest power cycles and it is usually covered in introductory courses on engineering thermodynamics (see, for example, Section 8.2 in \cite{moran2010fundamentals}). The Rankine cycle is made of four main components (i.e., compartments); a turbine, condenser, pump, and boiler. The aim of the cycle is to generate, via a turbine, mechanical work and produce electrical energy. The cycle is closed, that is, the working fluid circulates in loop within the four compartments. If we view the working fluid as a material, then we say that the Rankine cycle is a \emph{circular} life-cycle. Hence, we can think of the Rankine cycle, introduced in 1859 by the Scottish engineer William J. M. Rankine, as one of the first \emph{circular} systems. The compartmental diagram in Fig. \ref{fig:CombiningFigure} for a Rankine cycle would have 8 compartments connected in a loop, 4 of which are the pipes transporting the working fluid between the 4 main components mentioned above; in addition, it would have $Q = 1$ since the chemical composition of the working fluid is constant over the cycle.

\textbf{Topic 2 (CDT):} Along with circularity, the Rankine cycle has another important feature: its modeling begins by considering, in turn, the thermodynamics of each component (i.e., the turbine, condenser, pump, and boiler). Hence, the modeling is based on \emph{compartmental dynamical thermodynamics}, which conceives, as discussed in Section \ref{sub:CDT}, multi-machine systems as a set of thermodynamic compartments connected by material flows \cite{haddad2019dynamical}. Recall that the simplest analysis of a Rankine cycle is at steady state, i.e., when the time rate of change of energy $\dot{E} = 0$. Under the simplifying, but practically useful, assumption of steady state, the term ``dynamical'' in ``compartmental dynamical thermodynamics'' is removed since, at equilibrium, the dynamics of the system is not considered. 

\textbf{Topic 3 (RM):} As demonstrated in \cite{zocco2024towards}, the mechanics governing the dynamics of a robot can be derived from an energy balance. Hence, industrial manipulators can be considered as thermodynamic compartments. Recalling the analogy with the Rankine cycle, an industrial manipulator performing waste sorting is a thermodynamic compartment analogous to the turbine or the pump in the Rankine cycle. Now, given that a robotic manipulator is a thermodynamic compartment, we can assert that \emph{robotic manipulation} significantly affects the performance of the robot-compartment, and hence, the performance of the network of compartments as a whole. Therefore, an effective manipulation is essential for an effective recovery chain as depicted in Fig. \ref{fig:CombiningFigure}.     

\textbf{Topic 4 (DLV):} Consider again a manipulator as a thermodynamic compartment. If it is equipped with one or more cameras as in Fig. \ref{fig:CombiningFigure}, then these \emph{deep-learning vision} algorithms can process the images in real-time to improve the ability of the robot-compartment to handle waste materials or end-of-first-use products.   

\revision{In summary, Fig. 1 merges the four topics as follows. The rectangles are the control surfaces of the compartments (this is CDT); the robot depicted at the top in the recovery chain (green arrows) is a compartment that could sort or repair waste items using advanced manipulation and machine vision (this includes RM and DLV in the diagram). Finally, CE is included in the diagram since the material moves from a compartment to another in a circular fashion in order to reduce the two flows indicated with the red arrows. The literature review of each topic was covered in Section \ref{sec:4topics} in the following order: CE, CDT, RM, and DLV. The next section will review models and systems proposed in the literature that derive from mass balances and/or thermodynamic laws, and hence, they can be considered compartments $c^k_{i,j}$ to be connected as in Fig. \ref{fig:CombiningFigure} to simulate existing and hypothetical circular flows.}

\section{A BIBLIOTHECA OF COMPARTMENTS}\label{sec:library}
Since mass and energy balances are general natural principles, they have been used to develop many dynamical models to depict the dynamics of mechanical, chemical, and biological systems. In the context of this paper, these existing models are a library of thermodynamic compartments that can be added and connected as appropriate to design networks of thermodynamic compartments as the one depicted in Fig. \ref{fig:CombiningFigure}. They can also work as a starting point for developing custom compartments for a particular material flow design. Existing models of thermodynamic compartments are discussed below. 

In \cite{zocco2024towards} (specifically, in Proposition 1) the standard form of robot dynamics is derived from an energy balance, and thus, a manipulator can be considered a thermodynamic compartment. The robot manipulator can be used for waste sorting or product disassembly in reprocessing stages as shown in Fig. \ref{fig:CombiningFigure}. Similarly, a wheeled vehicle can be modeled as a thermodynamic compartment since its dynamics can be derived from the first law of thermodynamics (see \cite{liu2021garbage} and Proposition 1 in \cite{zocco2024towards}). In Fig. \ref{fig:CombiningFigure}, for the case where materials are moved on a truck (the case of a large vehicle) or if a robot is used to collect litter (the case of a small vehicle) \cite{almanzor2022autonomous}, a wheeled vehicle can be thought as a transportation compartment. 

Whenever the waste is located in areas with difficult access, an aerial drone may be a preferred mode for a transportation compartment \cite{gioioso2014flying}. The Lagrangian mechanics for an aerial drone is given in \cite{six2017kinematics}. The large amount of marine debris is becoming of increasing concern since microplastics can be ingested by fishes and, eventually, by humans \cite{lim2021microplastics}. To remove the debris materials and re-circulate them into production chains, underwater vehicles can be used \cite{zocco2023towards}. Hence, in this case, one of the transportation compartments in Fig. \ref{fig:CombiningFigure} would be an underwater vehicle, which recovers the waste materials from the ocean (the environment is indicated by $c^{n_c}_{n_c,n_c}$). The Lagrangian dynamics for an underwater vehicle is given in \cite{sagatun1991lagrangian}.

Thus far in this section, we covered four types of robot compartments, namely, manipulators, wheeled vehicles, aerial drones, and underwater vehicles. Their dynamics can be derived from energy or power balance equations (i.e., from the dynamical form of the first law of thermodynamics). There are of course other types of thermodynamic compartments whose models may include one or more mass balances. 

Compared to fossil fuels, bio-fuels are more carbon neutral because they are sourced from plantations, are more renewable, and can be produced locally to make a community self-sustained. Hence, bio-fuels are more aligned with circular economy principles than their fossil counterpart \cite{bandh2023biofuels}. A network of thermodynamic compartments for the biomass storage and production of bio-methane was designed in \cite{zocco2023thermodynamical}, where the transportation-compartment is a vehicle whose dynamics was covered above, while the model of the anaerobic digester follows from mass balances \cite{bernard2001dynamical,campos2019hybrid}.

The reuse of products or materials requires a preliminary washing step if they are infected. This is the case for devices used in healthcare such as surgical scissors, syringes, and inhalers. Hence, a circular flow of infected materials could involve a washing compartment along with transportation and robotic waste sorting. The thermodynamic model of a dishwasher heater was developed in \cite{mcdonald1989thermal}.

The study of material flows often involves the comparison between circular and linear systems, which are both shown in Fig. \ref{fig:CombiningFigure}; the former considers one or more material flow loops, whereas the latter sends material straight to incineration or the environment (indicated by $c^{n_c}_{n_c,n_c}$). For these comparative studies, the dynamics of an incinerator compartment is useful and it is provided in \cite{magnanelli2020dynamic} as a result of both mass and energy balances.

\revision{This section has reviewed existing models in the literature that fit within the proposed research framework. The next section will focus specifically on robots, which are a particular class of compartments that could play a significant role in a circular economy as depicted in Fig. \ref{fig:CombiningFigure}. The derivation of the standard form of robot dynamics from the first law of thermodynamics was demonstrated in \cite{zocco2024towards}, Proposition 1.}

\section{OPPORTUNITIES FOR ROBOTICS IN CIRCULAR ECONOMY}\label{sec:oppRobInCE}
Opportunities in application of robot technologies in a circular economy are wide-ranging and satisfy two key aspects connected with waste management: $i$) substitution of human activities in the case of elevated health risk due to the nature of the process or to the waste material properties; and $ii$) performance increment in waste sorting outputs or waste collection quality, avoiding high costs due to destination of improper material to the wrong waste treatment process in the new prospective of a circular economy.
Waste separated collection \cite{kumar2022design} and waste sorting were widely investigated with automation technologies focusing on the high level of waste flows to be managed and optimizing the high margins with respect to domestic waste management. The promising prospects of reducing management costs and increasing the effectiveness and efficiency of selection processes has stimulated the in-depth study of innovative robotization technologies \cite{sarc2019digitalisation}, in the face of a notable quantity of industrial automation solutions on a market scale with sustainable solutions applied to municipal solid waste \cite{lin2024development,gundupalli2017review}.

The transition from linear waste management to a circular economy requires a notable leap in quality in the performance of all phases of the process, in order to avoid effects of economic inefficiency due to the multiplication of transport and management costs due to flows affected by unsuitable materials with respect to the quality conditions required by recycling processes. From this perspective, robotization technologies can play a decisive role in the completion of automatic management of supply chains, in the final classification of flows destined for recycling in the face of reduced levels of unwanted or contaminated materials, and replacing processes that are still largely based on manual selection \cite{aschenbrenner2023robot}.

Further promising prospects for the application of robot technologies are: $i$) the disassembly processes of machinery and electronic equipment in general; and $ii$) the management of processes with high health risk. The first category of activities, which includes a notable number of research and development initiatives using robot technologies, is undoubtedly of considerable interest for the prospect of extending the concepts of circular economy to high economical value waste flows reducing the amount of precious materials today sent to disposal. The second sector of activity appears less investigated, but it is clear that interesting prospects (even if for waste flows of limited scope) can be applied to processes for the classification or identification of highly dangerous or radioactive materials erroneously present in transport processes or storage. In this case, robotization of on-site processes could reduce intervention times and the use of highly specialized personnel with notable benefits in all areas of logistics, which today are seriously damaged by the possible occurrence of incorrect deliveries in light of the controls now present in entrance to treatment plants. Similar application of on-site robotic processes are experienced and developed for the management of demolition and construction waste flows \cite{wang2020vision,chen2022robot}.

\revision{The next section will illustrate the application of the proposed framework to a case study of plastics including a robotic waste sorting compartment. This aligns with the current section, which identified waste sorting as a task to potentially automate with robots.}

\section{\revision{Framework Application to a Case Study and Challenges}}\label{sec:ForPlastics}
\subsection{\revision{Initial Application to Plastics Systems}}
\revision{This section covers the initial application of the proposed framework to design circular flows of plastics. Targeting plastics flows is motivated by the rising accumulation of plastic waste in natural environments \cite{lim2021microplastics}. This study considers also bio-plastics as they are made from renewable and carbon-sink feedstock, and hence, their use makes the communities more sustainable and resilient than with their synthetic counterpart \cite{BritishPF}. A project recently funded by the European Commission to enhance the circularity of the plastics industry is ReBioCycle \cite{BioRobWeb1,BioRobWeb2}, where robotic platforms are planned to be developed for waste sorting operations \cite{BioRobWeb2}.

We began the design of circular flows of plastics by looking at several existing and potential configurations of the supply-recovery chain. Adopting the nomenclature introduced in Fig. \ref{fig:CombiningFigure} and the methodology outlined by Zocco \emph{et al.} in \cite{zocco2023thermodynamical}, we conceptualize the five networks of thermodynamic compartments shown in Figs. \ref{fig:synthLin}-\ref{fig:bioRepair}, where $\beta_1$ is the raw material to make the considered plastic (synthetic or bio-based) and $\beta_2$ is either synthetic plastic (networks in Figs. \ref{fig:synthLin} and \ref{fig:synthCir}) or bio-plastic (networks in Figs. \ref{fig:bioCir}, \ref{fig:bioReuse}, and \ref{fig:bioRepair}). Since the compartmental networks result from a generalization of the Rankine cycle as explained in Section \ref{sec:combine}, Fig. \ref{fig:RankineAsTMN} shows the Rankine cycle, where $\beta_1$ is the working fluid. On the left-hand side, the cycle is depicted with the \emph{control surfaces} indicated by dashed lines as in Section 8.2 of \cite{moran2010fundamentals}, while on the left-hand side, the cycle is conceived as a network of compartments, with each compartment indicated by its control surface (a rectangle with solid line). The control surfaces indicate the boundaries of each compartment and similarly in Figs. \ref{fig:synthLin}-\ref{fig:bioRepair}. This way of representing sequences of interconnected compartments is called a ``compartmental diagram'' \cite{zocco2023thermodynamical}.      

Figs. \ref{fig:synthLin} and \ref{fig:synthCir} consider a linear system and a more circular system for synthetic plastic, respectively. The red arrows indicate the \emph{unsustainable flows} consisting of two types: raw materials extracted from non-renewable reservoirs entering the system and \emph{leaks of plastic} leaving the system. The \emph{return flows} into the supply chain are indicated with green arrows. As shown, both Fig. \ref{fig:synthLin} and Fig. \ref{fig:synthCir} have red flows in input and output, but in the former they are bigger because the latter retains a fraction of the plastic in the loop to meet the manufacturer demand $c^2_{2,2}$; the former has no return flows, while the latter has one return flow. The compartments that operate autonomously are depicted in yellow. The network in Fig. \ref{fig:synthCir} sorts the waste in an autonomous facility ($c^4_{4,4}$), which produces two streams: the multi-component products are sent to an autonomous disassembly facility to recover the plastic parts, which are then sent to the plastic recycler; and in contrast, plastic items (e.g., bottles, packaging) are sent directly to the recycler. Typically, a fraction of the plastic is not recyclable due to contamination or quality degradation; this causes the leak of plastic to $c^7_{7,7}$.

The remaining three subfigures of Fig. \ref{fig:PlasticNets} consider bio-plastic. Specifically, Fig. \ref{fig:bioCir} shows an existing chain, which recovers bio-plastic through the recycling of bio-plastic products. The network has one red flow instead of two because of the consideration of bio-plastic instead of synthetic plastic. In Fig. \ref{fig:bioReuse}, the bio-plastic components are disassembled from multi-component products and reused in products that differ from the initial ones due to the disassembly compartment ($c^6_{6,6}$), while in Fig. \ref{fig:bioRepair} bio-plastic components are reused in the original repaired products. Hence, while the former has two return flows to the manufacturer $c^2_{2,2}$ (the manufacturer uses the disassembled parts to make products), the latter has one return flow to the manufacturer and one to the retailer (the repaired products are ready to be sold again). Given that, to date, it is unlikely that multi-component products contain bio-plastic parts, Figs. \ref{fig:bioReuse} and \ref{fig:bioRepair} consider hypothetical compartmental configurations.             
\begin{figure*}
\centering
\begin{subfigure}{0.5\textwidth}
\includegraphics[width=\textwidth]{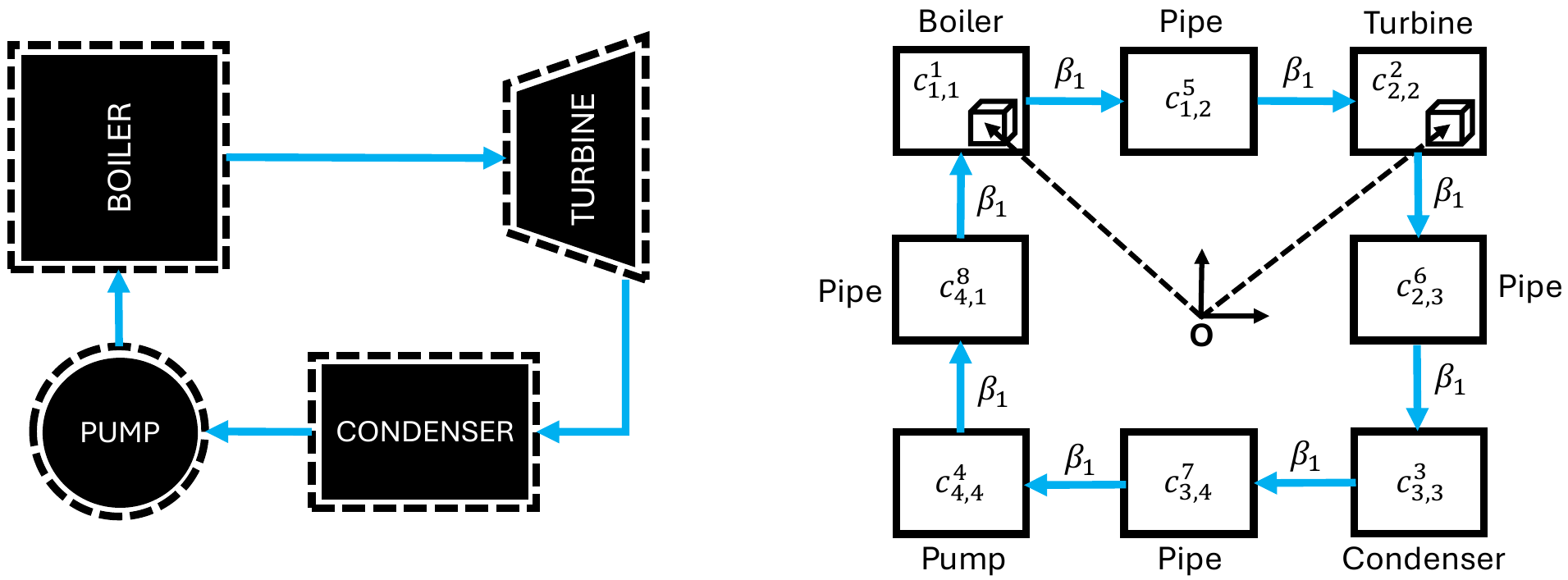}
\caption{\revision{Rankine cycle (left) conceived as a compartmental network (right)}}
\label{fig:RankineAsTMN}
\end{subfigure}
\begin{subfigure}{0.4\textwidth}
\centering
\includegraphics[width=\textwidth]{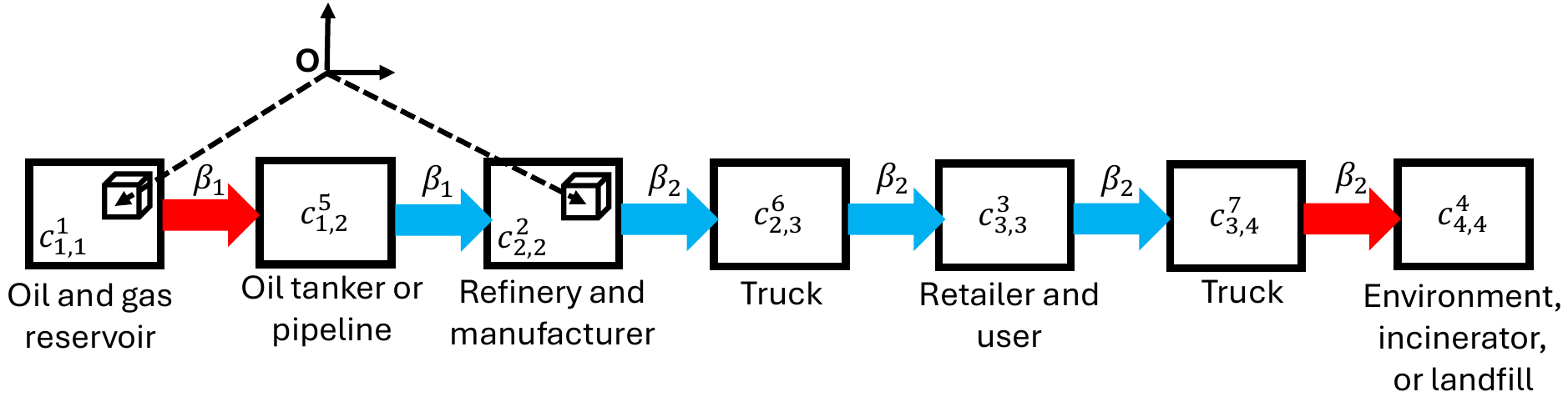}
\caption{\revision{Network of synthetic plastic (linear architecture)}}
\label{fig:synthLin}
\end{subfigure}
\begin{subfigure}{0.45\textwidth}
\centering
\includegraphics[width=\textwidth]{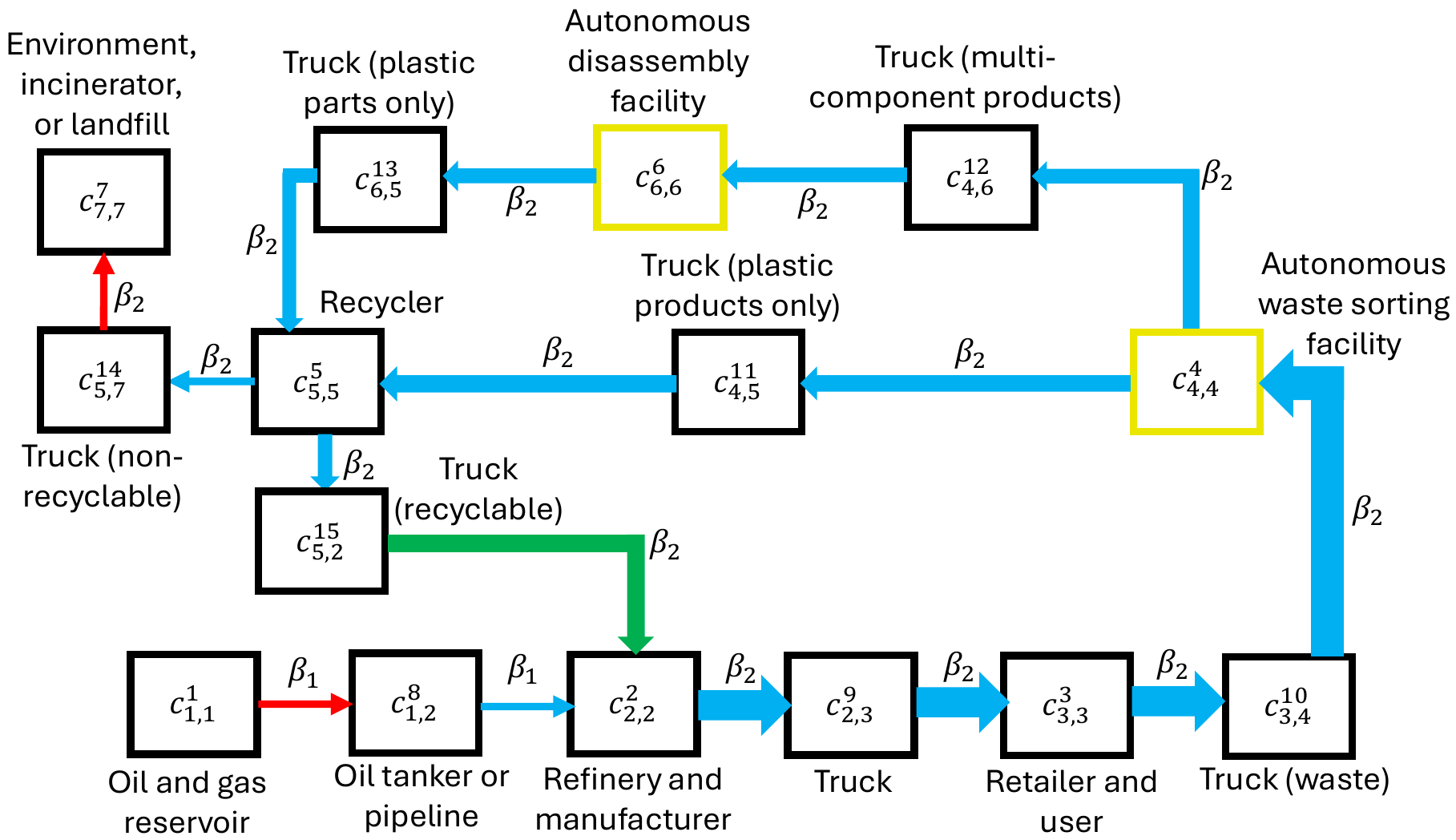}
\caption{\revision{Network of synthetic plastic (more circular architecture)}}
\label{fig:synthCir}
\end{subfigure}
\begin{subfigure}{0.45\textwidth}
\centering
\includegraphics[width=\textwidth]{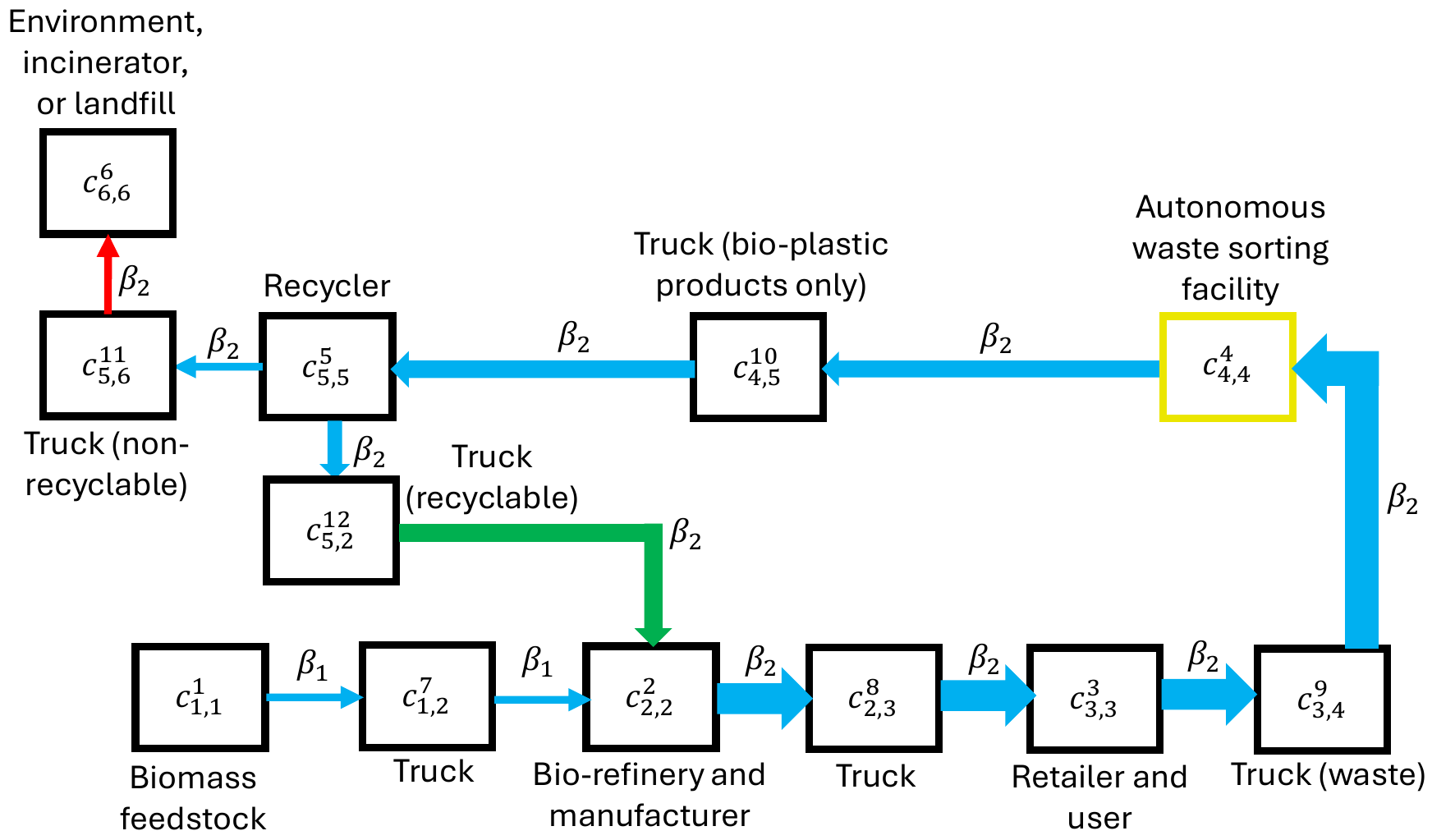}
\caption{\revision{Existing network of bio-plastic (no disassembly)}}
\label{fig:bioCir}
\end{subfigure}
\begin{subfigure}{0.45\textwidth}
\centering
\includegraphics[width=\textwidth]{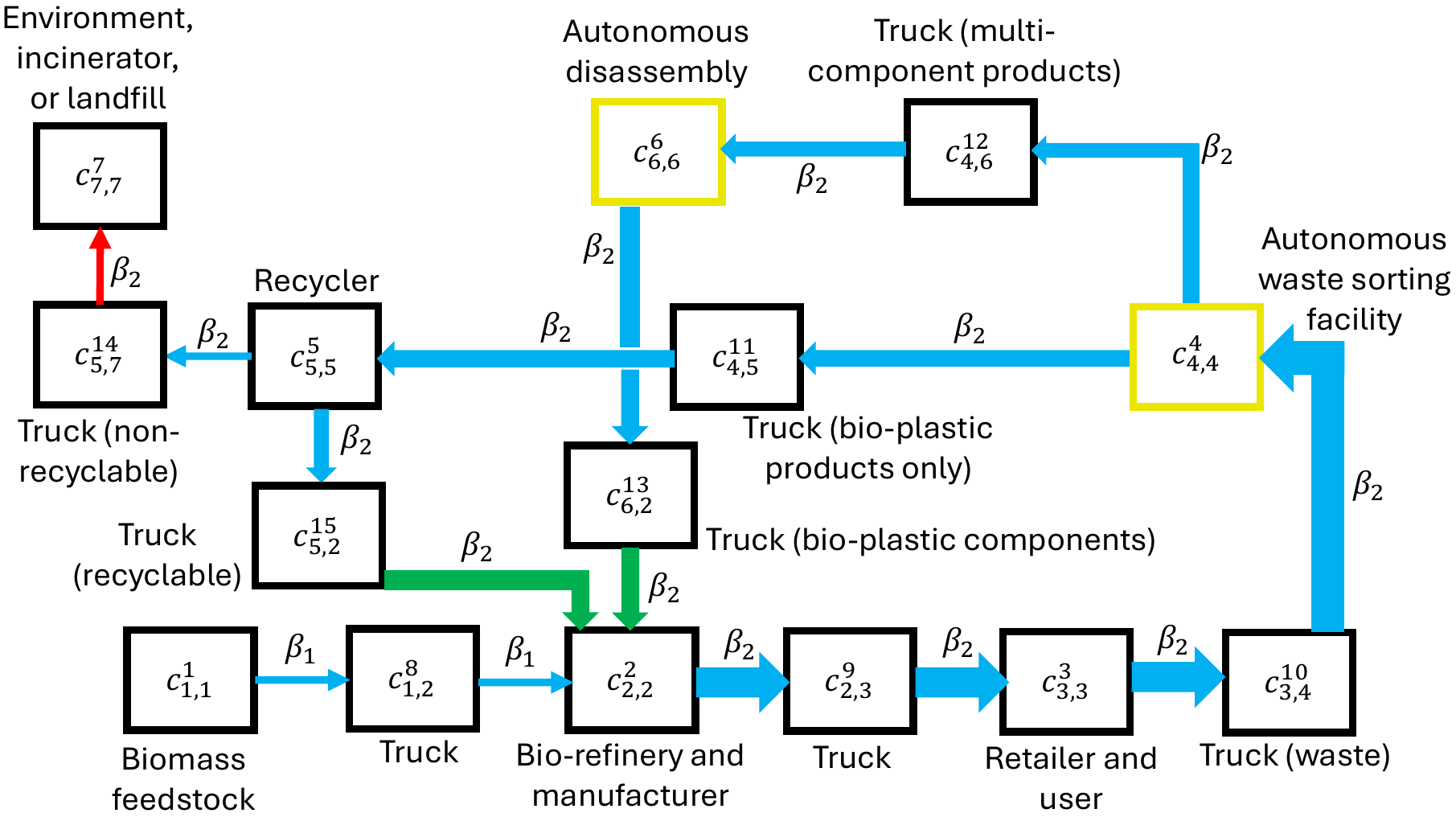}
\caption{\revision{Hypothetical net of bio-plastic (disassembly to reuse parts)}}
\label{fig:bioReuse}
\end{subfigure}
\begin{subfigure}{0.45\textwidth}
\centering
\includegraphics[width=\textwidth]{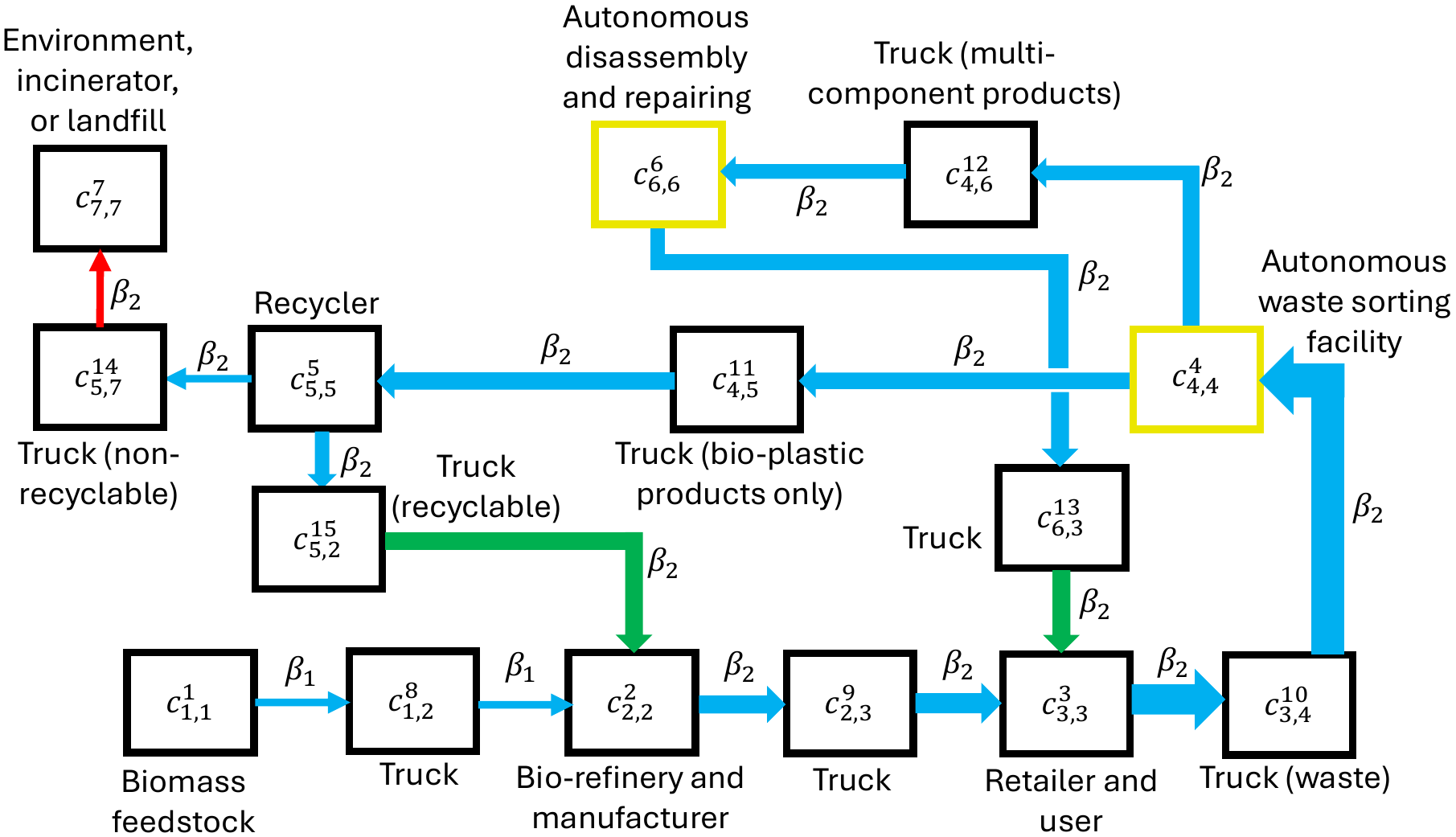}
\caption{\revision{Hypothetical net of bio-plastic (disassembly for repairing)}}
\label{fig:bioRepair}
\end{subfigure}
\caption{\revision{Initial application of the proposed framework to design plastic and bio-plastic flows. The framework is a generalization of the Rankine cycle (Fig. \ref{fig:RankineAsTMN}), where the working fluid is designed to be circular. At this initial stage, we focus on the sorting and the disassembly compartments performed via robotic manipulators controlled with deep RL (compartments in yellow).}}
\label{fig:PlasticNets}
\end{figure*}

Each network in Fig. \ref{fig:PlasticNets} can be represented by a set $\mathcal{N}$ of thermodynamic compartments \cite{zocco2023thermodynamical}. For example, the system in Fig. \ref{fig:synthCir} is represented by $\mathcal{N}_1$ as 
\begin{equation}
\begin{aligned}
\mathcal{N}_1 = \{c^1_{1,1},c^2_{2,2},c^3_{3,3},c^4_{4,4},c^5_{5,5},\\
c^6_{6,6},c^7_{7,7},c^8_{1,2},c^9_{2,3},c^{10}_{3,4},\\
c^{11}_{4,5},c^{12}_{4,6},c^{13}_{6,5},c^{14}_{5,7},c^{15}_{5,2}\}.
\end{aligned}
\end{equation}

Thus, the problem of designing circular flows can be stated as
\begin{equation}\label{eq:findOptimalNet}
\mathcal{N}^* = \argmin \,\,\, \dot{m}_\text{u}(\mathcal{N}),
\end{equation}
where $\dot{m}_\text{u}(\mathcal{N})$ is the sum of the unsustainable flows, i.e., the sum of the red flows, whose SI unit is kg/s. Equation (\ref{eq:findOptimalNet}) consists of finding the optimal network $\mathcal{N}^*$ that minimizes the unsustainable flows. Interventions on $\mathcal{N}$ to find $\mathcal{N}^*$ could be adding or removing compartments, modifying their interconnections, or changing their behaviour.
\begin{remark}
Note the analogy between (\ref{eq:findOptimalNet}) and the optimization of an artificial neural network $\mathcal{N}_{\textup{n}}$, which can be stated as 
\begin{equation}
\mathcal{N}^*_{\textup{n}} = \argmin \,\,\, L(\mathcal{N}_{\textup{n}}),
\end{equation}
where $L(\cdot)$ is a loss function \cite{Goodfellow-et-al-2016}. There is, however, a key difference between $\mathcal{N}_{\textup{n}}$ and $\mathcal{N}$: the constituent unit of the former is, in its simplest form, a linear combination of the input and the weights filtered by an activation function, i.e., the neuron; in contrast, the constituent unit of the latter is, in its simplest form, a first-order ordinary differential equation derived from a mass balance or a thermodynamic law, i.e., the compartment. 
\end{remark}

In this initial study, we focus on the sorting and disassembly of compartments being carried out by robots. The simulation results for a two-link revolute-revolute (RR) planar manipulator controlled with different RL agents are reported in Table \ref{tab:RLsuccRates}. The tested algorithms are the deep deterministic policy gradient (DDPG) \cite{lillicrap2015continuous}, the A2C \cite{A2C}, the proximal policy optimization (PPO) \cite{schulman2017proximal}, and the soft actor-critic (SAC) \cite{haarnoja2018soft}. For the DDPG algorithm, we used a multilayer perceptron as the policy model, a learning rate of 0.001, a batch size of 256, a soft update coefficient of 0.005, and a discount factor of 0.99. For the A2C algorithm, we used a multilayer perceptron as the policy model, a learning rate of 0.0007, a number of steps of 5, a discount factor of 0.99, an entropy coefficient of 0, a value function coefficient of 0.5, and the RMSprop optimizer. For the PPO algorithm, we used a multilayer perceptron as the policy model, a learning rate of 0.0003, a number of steps of 2048, a batch size of 64, a number of epochs of 10, a discount factor of 0.99, an entropy coefficient of 0, a value function coefficient of 0.5, and action noise exploration. Finally, for the SAC algorithm, we used a multilayer perceptron as the policy model, a learning rate of 0.0003, a batch size of 256, a soft update coefficient of 0.005, a discount factor of 0.99, a training frequency of 1, and an entropy regularization coefficient learned automatically.     

The simulation was performed using the Python library MuJoCo \cite{MuJoCo-Robots}. The action of the robot environment is a vector whose elements are the joint torques, while there are 11 observation variables including the distance between the tip of the second link and the target point. The target point is generated at a random position at the beginning of every simulation.   
\begin{table*}
\centering
\caption{\revision{Simulation performance of RL algorithms to control a two-link RR planar manipulator.}}
\label{tab:RLsuccRates}
\begin{tabular}{ccccc} 
\hline
Algorithm & Reward after training & Training time (min:sec) & Mean time per timestep (CPU; GPU) (sec) & Success rate\\ 
\hline
A2C & -10.99 & 4:56 & 0.00075; 0.0009 & 60.91\%\\
DDPG & -4.73 & 14:21 & 0.00058; 0.00071 & 93.37\%\\
PPO & -43.79 & 4:15 & 0.00073; 0.00092 & 41.83\%\\
SAC & -5.23 & 24:49 & 0.00079; 0.00102 & \textbf{95.97\%}\\
\hline 
\end{tabular}
\end{table*}
The success rate in Table \ref{tab:RLsuccRates} is calculated by considering a success whenever the tip of the second link reaches a distance from the target that is smaller than 4 cm and, simultaneously, the joint torques are smaller than 0.005 Nm (the reason for checking the torques along with the distance from the target is that the robot may reach the target and then, if the torques are high go beyond it, and thus failing the task; instead, the task is considered completed when the tip reaches the target and stops). The success rate is calculated over 10,000 simulations. The best algorithm is SAC followed by DDPG, although SAC requires also the longest time for training (24 min 49 sec). The next step will be to find a suitable RL algorithm to control a robot with at least 5 degrees of freedom performing pick-and-place in a waste sorting scenario, e.g., with a conveyor belt.

The simulations with the RR manipulator will also be used to analyse how the RL algorithm affects \emph{the whole network}. For example, the robot controlled via the SAC agent could sort $q$ items per hour, which results in $m_\text{r}$ tonnes of plastic sorted in 1 day. In this case, $m_\text{r}$ tonnes of plastic are sent to the recycling center whose recycling time takes $t_\text{r}$ hours. If the manufacturer requires $m_\text{r}$ tonnes in less than $t_\text{r}$ hours, then the manufacturer may buy some virgin plastic or order from another recycler. Moreover, the failure rate of SAC (4.03\%) indicates that some items will not be recycled resulting in a flow to the landfill that needs to be quantified and contained. The impact that the RL agent has on the supply-recovery chain can be captured only by looking at the whole network. This type of fundamental analysis regarding the material circularity is another important area of future work.

\subsection{\revision{Limitations and Challenges}}
\revision{Since the framework is based on the unique universal science of thermodynamics \cite{haddad2017thermodynamics}, it is very general. This generality renders the framework useful to designing any type of material and, given the network formalization, highly scalable. However, this generality may yield a lack of focus and the rise of confusion about its purpose. By proposing the framework within a circular economy setting, we hope to provide the required clarity to the reader on the final goal that motivated its conceptualization. Another main challenge will be to create and maintain a two-way exchange of design ideas of the networks such as those in Fig. \ref{fig:PlasticNets} between industry, with its practicality, and researchers, with their creativity. Without the former, the innovative ideas would remain in papers; without the latter, the possible pathways for a transition towards a circular society would be more difficult to envision. 

On a more technical level, the first main challenge in implementing the framework to real scenarios is that the number of compartments to deal with could be large even for small supply-recovery chains because the materials and the product life-cycle stages in a modern society are many and highly dynamic. The second main challenge is to translate any network or compartment alterations into their impact on the unsustainable flows as compactly expressed by Equation (\ref{eq:findOptimalNet}). Specifically, if a compartment is, compartment-wise, far from any unsustainable flows, then the quantification of the impact of its dynamics on the unsustainable flows requires to back- and forward-propagate the variation to the following and proceeding compartments that separate it from the unsustainable flows. For example, the performance of the RL controller in the waste sorting facility ($c^4_{4,4}$) affects directly the neighboring compartments ($c^{12}_{4,6}$, $c^{11}_{4,5}$, and $c^{10}_{3,4}$). However, to quantify its impact on the red arrows it is necessary to propagate its impact on the intermediate compartments one at a time both forward ($c^{12}_{4,6}$, $c^{6}_{6,6}$, $c^{13}_{6,5}$, etc.) and backward ($c^{10}_{3,4}$, $c^{3}_{3,3}$, etc.) until the red flows are reached.}
}

\section{CONCLUSION}\label{sec:concl}
\revision{In this paper, we merged deep-learning vision, compartmental dynamical thermodynamics, and robotic manipulation with circular economy.  
Deep-learning vision is advancing rapidly due to performance improvements via neural networks, but its use for the automation of recovery tasks such as waste sorting and disassembly remains relatively undeveloped.} The adaptability and scalability of deep-learning vision can enable highly-flexible waste sorting facilities to process very different types of materials and products. This is currently not feasible with traditional automated methods, which can process only particular classes of items. In addition, deep-learning vision can be used for wide-area autonomous materials monitoring to improve natural resources management (as detailed in \cite{zocco2024synchronized}). 

Another key ingredient for highly-flexible waste sorting systems are advanced robotic manipulators capable of handling items with shapes difficult to predict such as those of end-of-life products. However, advanced manipulation such as the one based on soft components poses new challenges for planning and control of robot motions. To align the grippers with circular economy principles, the use of bio- or bio-degradable materials in their designs is an option worth exploring. The benefits of an effective autonomous waste sorting system include a reduction in contamination risks by human operators, a reduction in operating costs, and an increase of the rate of material separation for reuse. An inefficient separation causes a higher fraction of material being sent for incineration or, illegally or not, to the environment, thus becoming litter and marine debris. 

The integration of deep-learning vision and robotic manipulation can also be used in disassembly operations to repair faulty items. Given that, to date, the technological level in manufacturing is significantly more advanced than in re-manufacturing, there are significant opportunities in transferring the knowledge from the former to the latter. This will certainly be a challenging task since the geometries and conditions of end-of-life products are much less predictable than the ones of new items. Disassembly for the recovery of precious metals and critical raw materials is particularly promising due to the economic value of these natural resources, which justifies the costs of robotic systems. 

The term ``circular'' itself implies a systemic perspective that considers the material life-cycle from extraction to its reuse and disposal. Hence, this paper proposes compartmental dynamical thermodynamics to establish rigorous physics-based theoretical foundations of circular flow designs. Systems such as robots, transportation systems, and incinerators, obey the laws of thermodynamics, and hence, their interactions form a network of thermodynamic compartments. The thermodynamic compartments can be added, modified, or removed at the design stage of the compartmental network in order to achieve a circular material flow. As we showed in this paper, a theoretically-coherent physics-based framework underpinning the design of circular flow systems is missing in the literature and it is needed to speed-up the implementation of circularity at the local and global scales.

\revision{The application of the proposed framework to the case study of plastics systems highlighted several points. Firstly, the framework provides a clear and intuitive representation of supply-recovery chains, especially for those with an industrial and thermal engineering expertise since the framework results from the generalization of the design of thermodynamic cycles such as the Rankine cycle. Secondly, the framework is flexible and scalable, and hence, it enables the consideration of existing network configurations along with simulating hypothetical architectures as needed in what-if analysis. Thirdly, compared to existing frameworks for CE, this framework can coherently integrate robotics and autonomous systems since it is predicated on dynamical systems modeling rather than data analysis as in traditional approaches to CE, e.g., MFA \cite{zocco2023thermodynamical}. Fourthly, RL controllers have great potential for making robotic manipulators as flexible as humans, but currently it is an unexplored area in the literature for waste sorting and disassembly operations. As a result, its development requires to work not only on finding and tuning a suitable RL algorithm, but also creating the RL simulation and testing environment for CE applications. Fifthly, the conceptualization of CE as an optimization problem as in Equation (\ref{eq:findOptimalNet}) highlights that variations occurring within a single compartment need to be propagated to the other compartments until the unsustainable flows are reached in order to quantify the impact of the compartmental variation on the network circularity. This is because there is a strong interrelationship between compartment-level dynamics and the overall network circularity. Sixthly, Equation (\ref{eq:findOptimalNet}) also has highlighted the similarity and the difference between compartmental and artificial neural networks. While both are designed as a result of an optimization problem, their constituent units are fundamentally different; with compartments being employed in the former and the neurons in the latter.

Future work will consist on developing RL environments and algorithms for CE applications such as robotic waste sorting and disassembly; this could be followed by their experimentation on real manipulators. Another future research direction could be to simulate the plastics systems discussed in the case study in order to analyse existing and hypothetical configurations in collaboration with industry and supply-chain practitioners.}

\section*{ACKNOWLEDGMENT}
F. Zocco sincerely thanks the Department of Information Engineering and Mathematics of University of Siena (Italy) for hosting him for this research \revision{and Mohd Shoaib at Nottingham Trent University (UK) for the valuable discussions on reinforcement learning.}

\bibliographystyle{IEEEtran}
\bibliography{bibliography}

\begin{thebibliography}{100}
\providecommand{\url}[1]{#1}
\csname url@samestyle\endcsname
\providecommand{\newblock}{\relax}
\providecommand{\bibinfo}[2]{#2}
\providecommand{\BIBentrySTDinterwordspacing}{\spaceskip=0pt\relax}
\providecommand{\BIBentryALTinterwordstretchfactor}{4}
\providecommand{\BIBentryALTinterwordspacing}{\spaceskip=\fontdimen2\font plus
\BIBentryALTinterwordstretchfactor\fontdimen3\font minus
  \fontdimen4\font\relax}
\providecommand{\BIBforeignlanguage}[2]{{%
\expandafter\ifx\csname l@#1\endcsname\relax
\typeout{** WARNING: IEEEtran.bst: No hyphenation pattern has been}%
\typeout{** loaded for the language `#1'. Using the pattern for}%
\typeout{** the default language instead.}%
\else
\language=\csname l@#1\endcsname
\fi
#2}}
\providecommand{\BIBdecl}{\relax}
\BIBdecl

\bibitem{CRMall}
CRM\hspace{3pt}Alliance, 2024, \, webpage:
  \url{https://www.crmalliance.eu/critical-raw-materials}; last access: 9 April
  2024.

\bibitem{CRM-EU}
European\hspace{3pt}Commission, 2023, \, webpage:
  \url{https://single-market-economy.ec.europa.eu/sectors/raw-materials/areas-specific-interest/critical-raw-materials\_en};
  last access: 9 April 2024.

\bibitem{CRM-US}
Department\hspace{3pt}of\hspace{3pt}Energy, 2023, \, webpage:
  \url{https://www.energy.gov/cmm/what-are-critical-materials-and-critical-minerals};
  last access: 9 April 2024.

\bibitem{NASA-CO2}
NASA, 2024, \, webpage:
  \url{https://climate.nasa.gov/vital-signs/carbon-dioxide/?intent=111}; last
  access: 9 April 2024.

\bibitem{NASA-effects}
------, 2024, \, webpage:
  \url{https://science.nasa.gov/climate-change/effects/}; last access: 9 April
  2024.

\bibitem{UN2024}
United\hspace{3pt}Nations, ``Beyond an age of waste,'' 2024, \, available at:
  \url{https://wedocs.unep.org/bitstream/handle/20.500.11822/44939/global\_waste\_management\_outlook\_2024.pdf?sequence=3};
  last access: 9 April 2024.

\bibitem{OECD2023}
OECD, ``Circular economy - waste and materials,'' 2023, \, available at:
  \url{https://www.oecd.org/environment/environment-at-a-glance/Environment%20at%20a%20Glance%20Indicators%20Circular%20economy%20waste%20and%20materials%20Q2.pdf};
  last access: 9 April 2024.

\bibitem{EllenMAF}
Ellen\hspace{3pt}MacArthur\hspace{3pt}Foundation, ``What is a circular
  economy?'' 2024, \, webpage:
  \url{https://www.ellenmacarthurfoundation.org/topics/circular-economy-introduction/overview};
  last access: 9 April 2024.

\bibitem{russell2023value}
J.~D. Russell and N.~Z. Nasr, ``Value-retained vs. impacts avoided: {The}
  differentiated contributions of remanufacturing, refurbishment, repair, and
  reuse within a circular economy,'' \emph{Journal of Remanufacturing},
  vol.~13, no.~1, pp. 25--51, 2023.

\bibitem{mederake2023without}
L.~Mederake, ``Without a debate on sufficiency, a circular plastics economy
  will remain an illusion,'' \emph{Circular Economy and Sustainability},
  vol.~3, no.~3, pp. 1425--1439, 2023.

\bibitem{CEPSonCE}
Centre\hspace{3pt}for\hspace{3pt}European\hspace{3pt}Policy\hspace{3pt}Studies\hspace{3pt}(CEPS),
  ``Barriers and enablers for implementing circular economy business models,''
  2021, \, available at:
  \url{https://cdn.ceps.eu/wp-content/uploads/2021/10/RR2021-01_Barriers-and-enablers-for-implementing-circular-economy-business-models.pdf};
  last access: 27 September 2024.

\bibitem{kirchherr2019towards}
J.~Kirchherr and L.~Piscicelli, ``Towards an education for the circular economy
  {(ECE): Five} teaching principles and a case study,'' \emph{Resources,
  Conservation and Recycling}, vol. 150, p. 104406, 2019.

\bibitem{mesa2021towards}
J.~A. Mesa and I.~Esparragoza, ``Towards the implementation of circular economy
  in engineering education: A systematic review,'' in \emph{2021 IEEE Frontiers
  in Education Conference (FIE)}.\hskip 1em plus 0.5em minus 0.4em\relax IEEE,
  2021, pp. 1--8.

\bibitem{giannoccaro2021features}
I.~Giannoccaro, G.~Ceccarelli, and L.~Fraccascia, ``Features of the higher
  education for the circular economy: {The case of Italy},''
  \emph{Sustainability}, vol.~13, no.~20, p. 11338, 2021.

\bibitem{bakshi2022sustainability}
B.~R. Bakshi and J.~A. Paulson, ``Sustainability and industry 4.0: Obstacles
  and opportunities,'' in \emph{2022 American Control Conference (ACC)}.\hskip
  1em plus 0.5em minus 0.4em\relax IEEE, 2022, pp. 2449--2460.

\bibitem{velenturf2021principles}
A.~P. Velenturf and P.~Purnell, ``Principles for a sustainable circular
  economy,'' \emph{Sustainable Production and Consumption}, vol.~27, pp.
  1437--1457, 2021.

\bibitem{kirchherr2017conceptualizing}
J.~Kirchherr, D.~Reike, and M.~Hekkert, ``Conceptualizing the circular economy:
  An analysis of 114 definitions,'' \emph{Resources, Conservation and
  Recycling}, vol. 127, pp. 221--232, 2017.

\bibitem{moraga2019circular}
G.~Moraga, S.~Huysveld, F.~Mathieux, G.~A. Blengini, L.~Alaerts, K.~Van~Acker,
  S.~De~Meester, and J.~Dewulf, ``Circular economy indicators: What do they
  measure?'' \emph{Resources, Conservation and Recycling}, vol. 146, pp.
  452--461, 2019.

\bibitem{saidani2019taxonomy}
M.~Saidani, B.~Yannou, Y.~Leroy, F.~Cluzel, and A.~Kendall, ``A taxonomy of
  circular economy indicators,'' \emph{Journal of Cleaner Production}, vol.
  207, pp. 542--559, 2019.

\bibitem{zocco2022circularity}
F.~Zocco, B.~Smyth, and P.~Sopasakis, ``Circularity of thermodynamical material
  networks: Indicators, examples, and algorithms,'' \emph{arXiv preprint
  arXiv:2209.15051}, 2022.

\bibitem{de2021systematic}
A.~De~Pascale, R.~Arbolino, K.~Szopik-Depczy{\'n}ska, M.~Limosani, and
  G.~Ioppolo, ``A systematic review for measuring circular economy: The 61
  indicators,'' \emph{Journal of Cleaner Production}, vol. 281, p. 124942,
  2021.

\bibitem{rosa2020assessing}
P.~Rosa, C.~Sassanelli, A.~Urbinati, D.~Chiaroni, and S.~Terzi, ``Assessing
  relations between {Circular Economy} and {Industry} 4.0: A systematic
  literature review,'' \emph{International Journal of Production Research},
  vol.~58, no.~6, pp. 1662--1687, 2020.

\bibitem{laskurain2021exploring}
I.~Laskurain-Iturbe, G.~Arana-Land{\'\i}n, B.~Landeta-Manzano, and
  N.~Uriarte-Gallastegi, ``Exploring the influence of industry 4.0 technologies
  on the circular economy,'' \emph{Journal of Cleaner Production}, vol. 321, p.
  128944, 2021.

\bibitem{kumar2024barriers}
A.~Kumar, S.~K. Mangla, and P.~Kumar, ``Barriers for adoption of {Industry 4.0}
  in sustainable food supply chain: A circular economy perspective,''
  \emph{International Journal of Productivity and Performance Management},
  vol.~73, no.~2, pp. 385--411, 2024.

\bibitem{leising2018circular}
E.~Leising, J.~Quist, and N.~Bocken, ``Circular economy in the building sector:
  Three cases and a collaboration tool,'' \emph{Journal of Cleaner production},
  vol. 176, pp. 976--989, 2018.

\bibitem{howard2022going}
M.~Howard, X.~Yan, N.~Mustafee, F.~Charnley, S.~B{\"o}hm, and S.~Pascucci,
  ``Going beyond waste reduction: Exploring tools and methods for circular
  economy adoption in small-medium enterprises,'' \emph{Resources, Conservation
  and Recycling}, vol. 182, p. 106345, 2022.

\bibitem{brunner2016handbook}
P.~H. Brunner and H.~Rechberger, \emph{{Handbook of Material Flow Analysis: For
  Environmental, Resource, and Waste Engineers}}.\hskip 1em plus 0.5em minus
  0.4em\relax CRC press, 2016.

\bibitem{CSIRO-MFA}
Commonwealth\hspace{3pt}Scientific\hspace{3pt}and\hspace{3pt}Research\hspace{3pt}Organization\hspace{3pt}(CSIRO),
  ``Material flow analysis to progress to a circular economy,'' 2024, \,
  webpage:
  \url{https://research.csiro.au/circulareconomy/material-flow-report/}; last
  access: 26 September 2024.

\bibitem{EU-MFA}
Eurostat, ``Circular economy -- material flows,'' 2023, \, webpage:
  \url{https://ec.europa.eu/eurostat/statistics-explained/index.php?title=Circular_economy_-_material_flows#Sankey_diagram_of_material_flows};
  last access: 26 September 2024.

\bibitem{cullen2022material}
J.~M. Cullen and D.~R. Cooper, ``Material flows and efficiency,'' \emph{Annual
  Review of Materials Research}, vol.~52, no.~1, pp. 525--559, 2022.

\bibitem{zocco2023thermodynamical}
F.~Zocco, P.~Sopasakis, B.~Smyth, and W.~M. Haddad, ``Thermodynamical material
  networks for modeling, planning, and control of circular material flows,''
  \emph{International Journal of Sustainable Engineering}, vol.~16, no.~1, pp.
  1--14, 2023.

\bibitem{haddad2009thermodynamics}
W.~M. Haddad, V.~Chellaboina, and S.~G. Nersesov, \emph{Thermodynamics: {A
  Dynamical Systems Approach}}.\hskip 1em plus 0.5em minus 0.4em\relax
  Princeton University Press, 2005.

\bibitem{haddad2013unification}
W.~M. Haddad, ``A unification between dynamical system theory and
  thermodynamics involving an energy, mass, and entropy state space
  formalism,'' \emph{Entropy}, vol.~15, no.~5, pp. 1821--1846, 2013.

\bibitem{haddad2017thermodynamics}
------, ``Thermodynamics: The unique universal science,'' \emph{Entropy},
  vol.~19, no. 621, pp. 1--70, 2017.

\bibitem{haddad2020condensed}
------, ``Condensed matter physics, hybrid energy and entropy principles, and
  the hybrid first and second laws of thermodynamics,'' \emph{Communications in
  Nonlinear Science and Numerical Simulation}, vol.~83, p. 105096, 2020.

\bibitem{haddad2019dynamical}
------, \emph{A Dynamical Systems Theory of Thermodynamics}.\hskip 1em plus
  0.5em minus 0.4em\relax Princeton University Press, 2019.

\bibitem{kiyokawa2022challenges}
T.~Kiyokawa, J.~Takamatsu, and S.~Koyanaka, ``Challenges for future robotic
  sorters of mixed industrial waste: A survey,'' \emph{IEEE Transactions on
  Automation Science and Engineering}, vol.~21, no.~1, pp. 1023--1040, 2022.

\bibitem{billard2019trends}
A.~Billard and D.~Kragic, ``Trends and challenges in robot manipulation,''
  \emph{Science}, vol. 364, no. 6446, pp. 1--8 (eaat8414), 2019.

\bibitem{prattichizzo2016grasping}
D.~Prattichizzo and J.~C. Trinkle, ``Grasping,'' \emph{Springer Handbook of
  Robotics}, pp. 955--988, 2016.

\bibitem{zaidi2021actuation}
S.~Zaidi, M.~Maselli, C.~Laschi, and M.~Cianchetti, ``Actuation technologies
  for soft robot grippers and manipulators: A review,'' \emph{Current Robotics
  Reports}, vol.~2, no.~3, pp. 355--369, 2021.

\bibitem{marullo2020mag}
S.~Marullo, S.~Bartoccini, G.~Salvietti, M.~Z. Iqbal, and D.~Prattichizzo,
  ``The mag-gripper: A soft-rigid gripper augmented with an electromagnet to
  precisely handle clothes,'' \emph{IEEE Robotics and Automation Letters},
  vol.~5, no.~4, pp. 6591--6598, 2020.

\bibitem{li2024grasp}
Y.~Li, B.~Liu, Y.~Geng, P.~Li, Y.~Yang, Y.~Zhu, T.~Liu, and S.~Huang, ``Grasp
  multiple objects with one hand,'' \emph{IEEE Robotics and Automation
  Letters}, vol.~9, no.~5, pp. 4027--4034, 2024.

\bibitem{qu2023recent}
J.~Qu, B.~Mao, Z.~Li, Y.~Xu, K.~Zhou, X.~Cao, Q.~Fan, M.~Xu, B.~Liang, H.~Liu
  \emph{et~al.}, ``Recent progress in advanced tactile sensing technologies for
  soft grippers,'' \emph{Advanced Functional Materials}, vol.~33, no.~41, p.
  2306249, 2023.

\bibitem{eppner2015exploitation}
C.~Eppner, R.~Deimel, J.~Alvarez-Ruiz, M.~Maertens, and O.~Brock,
  ``Exploitation of environmental constraints in human and robotic grasping,''
  \emph{The International Journal of Robotics Research}, vol.~34, no.~7, pp.
  1021--1038, 2015.

\bibitem{pozzi2020hand}
M.~Pozzi, S.~Marullo, G.~Salvietti, J.~Bimbo, M.~Malvezzi, and D.~Prattichizzo,
  ``Hand closure model for planning top grasps with soft robotic hands,''
  \emph{The International Journal of Robotics Research}, vol.~39, no.~14, pp.
  1706--1723, 2020.

\bibitem{bo2022automated}
V.~Bo, E.~Turco, M.~Pozzi, M.~Malvezzi, and D.~Prattichizzo, ``Automated design
  of embedded constraints for soft hands enabling new grasp strategies,''
  \emph{IEEE Robotics and Automation Letters}, vol.~7, no.~4, pp.
  11\,346--11\,353, 2022.

\bibitem{ku2021deep}
Y.~Ku, J.~Yang, H.~Fang, W.~Xiao, and J.~Zhuang, ``Deep learning of grasping
  detection for a robot used in sorting construction and demolition waste,''
  \emph{Journal of Material Cycles and Waste Management}, vol.~23, pp. 84--95,
  2021.

\bibitem{papadakis2020use}
E.~Papadakis, F.~Raptopoulos, M.~Koskinopoulou, and M.~Maniadakis, ``On the use
  of vacuum technology for applied robotic systems,'' in \emph{2020 6th
  International Conference on Mechatronics and Robotics Engineering}.\hskip 1em
  plus 0.5em minus 0.4em\relax IEEE, 2020, pp. 73--77.

\bibitem{9216746}
F.~Raptopoulos, M.~Koskinopoulou, and M.~Maniadakis, ``Robotic pick-and-toss
  facilitates urban waste sorting,'' in \emph{2020 IEEE 16th International
  Conference on Automation Science and Engineering}, 2020, pp. 1149--1154.

\bibitem{koskinopoulou2021robotic}
M.~Koskinopoulou, F.~Raptopoulos, G.~Papadopoulos, N.~Mavrakis, and
  M.~Maniadakis, ``Robotic waste sorting technology: Toward a vision-based
  categorization system for the industrial robotic separation of recyclable
  waste,'' \emph{IEEE Robotics \& Automation Magazine}, vol.~28, no.~2, pp.
  50--60, 2021.

\bibitem{um2023fast}
S.~Um, K.-S. Kim, and S.~Kim, ``Fast suction-grasp-difficulty estimation for
  high throughput plastic-waste sorting,'' \emph{Journal of Mechanical Science
  and Technology}, vol.~37, no.~2, pp. 955--964, 2023.

\bibitem{9551603}
------, ``Suction point selection algorithm based on point cloud for plastic
  waste sorting,'' in \emph{2021 IEEE 17th International Conference on
  Automation Science and Engineering}, 2021, pp. 60--65.

\bibitem{engelen2023high}
B.~Engelen, D.~De~Marelle, J.~Peeters, and K.~Kellens, ``High airflow vertical
  conveying gripper for robotic sorting of shredded metal residues,''
  \emph{Procedia CIRP}, vol. 116, pp. 396--401, 2023.

\bibitem{rithvik2023soft}
S.~Rithvik, V.~Rai, S.~Dornal, J.~Deepak, and B.~Pramod, ``Soft robotics in
  waste management,'' \emph{Self-Powered Cyber Physical Systems}, pp. 341--348,
  2023.

\bibitem{achilli2023multibody}
G.~M. Achilli, S.~Logozzo, and M.~Malvezzi, ``Multibody simulation of an
  underactuated gripper for sustainable waste sorting,'' in \emph{International
  Workshop IFToMM for Sustainable Development Goals}.\hskip 1em plus 0.5em
  minus 0.4em\relax Springer, 2023, pp. 476--483.

\bibitem{9530533}
T.~Kiyokawa, H.~Katayama, Y.~Tatsuta, J.~Takamatsu, and T.~Ogasawara, ``Robotic
  waste sorter with agile manipulation and quickly trainable detector,''
  \emph{IEEE Access}, vol.~9, pp. 124\,616--124\,631, 2021.

\bibitem{achilli2023underactuated}
G.~M. Achilli, S.~Logozzo, M.~Malvezzi, and M.~C. Valigi, ``Underactuated
  embedded constraints gripper for grasping in toxic environments,'' \emph{SN
  Applied Sciences}, vol.~5, no.~4, p.~96, 2023.

\bibitem{crooks2016fin}
W.~Crooks, G.~Vukasin, M.~O’Sullivan, W.~Messner, and C.~Rogers, ``Fin
  ray{\textregistered} effect inspired soft robotic gripper: From the robosoft
  grand challenge toward optimization,'' \emph{Frontiers in Robotics and AI},
  vol.~3, p.~70, 2016.

\bibitem{almanzor2022autonomous}
E.~Almanzor, N.~R. Anvo, T.~G. Thuruthel, and F.~Iida, ``Autonomous detection
  and sorting of litter using deep learning and soft robotic grippers,''
  \emph{Frontiers in Robotics and AI}, vol.~9, p. 1064853, 2022.

\bibitem{sadeghian2022vision}
R.~Sadeghian, S.~Shahin, and S.~Sareh, ``Vision-based self-adaptive gripping in
  a trimodal robotic sorting end-effector,'' \emph{IEEE Robotics and Automation
  Letters}, vol.~7, no.~2, pp. 2124--2131, 2022.

\bibitem{9981625}
J.~Buzzatto, M.~Shahmohammadi, J.~Liang, F.~Sanches, S.~Matsunaga,
  R.~Haraguchi, T.~Mariyama, B.~MacDonald, and M.~Liarokapis, ``Soft,
  multi-layer, disposable, kirigami based robotic grippers: On handling of
  delicate, contaminated, and everyday objects,'' in \emph{2022 IEEE/RSJ
  International Conference on Intelligent Robots and Systems}, 2022, pp.
  5440--5447.

\bibitem{bonello2017exploratory}
D.~Bonello, M.~A. Saliba, and K.~P. Camilleri, ``An exploratory study on the
  automated sorting of commingled recyclable domestic waste,'' \emph{Procedia
  Manufacturing}, vol.~11, pp. 686--694, 2017.

\bibitem{kang2024enhancing}
H.~Kang, S.~Im, J.~Jo, and M.-S. Kang, ``Enhancing recycling efficiency: A
  rapid glass bottle sorting gripper,'' \emph{Robotics and Autonomous Systems},
  vol. 174, p. 104647, 2024.

\bibitem{daneshmand2023industry}
M.~Daneshmand, F.~Noroozi, C.~Corneanu, F.~Mafakheri, and P.~Fiorini,
  ``Industry 4.0 and prospects of circular economy: A survey of robotic
  assembly and disassembly,'' \emph{The International Journal of Advanced
  Manufacturing Technology}, vol. 124, no.~9, pp. 2973--3000, 2023.

\bibitem{tan2021hybrid}
W.~J. Tan, C.~M.~M. Chin, A.~Garg, and L.~Gao, ``A hybrid disassembly framework
  for disassembly of electric vehicle batteries,'' \emph{International Journal
  of Energy Research}, vol.~45, no.~5, pp. 8073--8082, 2021.

\bibitem{kay2022robotic}
I.~Kay, S.~Farhad, A.~Mahajan, R.~Esmaeeli, and S.~R. Hashemi, ``Robotic
  disassembly of electric vehicles’ battery modules for recycling,''
  \emph{Energies}, vol.~15, no.~13, p. 4856, 2022.

\bibitem{vongbunyong2015disassembly}
S.~Vongbunyong, W.~H. Chen, S.~Vongbunyong, and W.~H. Chen, \emph{Disassembly
  Automation}.\hskip 1em plus 0.5em minus 0.4em\relax Springer, 2015.

\bibitem{foo2022challenges}
G.~Foo, S.~Kara, and M.~Pagnucco, ``Challenges of robotic disassembly in
  practice,'' \emph{Procedia CIRP}, vol. 105, pp. 513--518, 2022.

\bibitem{farhan2021autonomous}
A.~S. Farhan, A.~M. Bassiouny, Y.~T. Afif, A.~A. Gamil, M.~A. Alsheikh, A.~M.
  Kamal, K.~S. Elenany, M.~A. Bahour, M.~I. Awad, and S.~A. Maged, ``Autonomous
  non-destructive assembly/disassembly of electronic components using a robotic
  arm,'' in \emph{2021 16th International Conference on Computer Engineering
  and Systems}.\hskip 1em plus 0.5em minus 0.4em\relax IEEE, 2021, pp. 1--7.

\bibitem{marconi2019feasibility}
M.~Marconi, G.~Palmieri, M.~Callegari, and M.~Germani, ``Feasibility study and
  design of an automatic system for electronic components disassembly,''
  \emph{Journal of Manufacturing Science and Engineering}, vol. 141, no.~2, p.
  021011, 2019.

\bibitem{hathaway2023towards}
J.~Hathaway, A.~Shaarawy, C.~Akdeniz, A.~Aflakian, R.~Stolkin, and
  A.~Rastegarpanah, ``Towards reuse and recycling of lithium-ion batteries:
  Tele-robotics for disassembly of electric vehicle batteries,''
  \emph{Frontiers in Robotics and AI}, vol.~10, p. 1179296, 2023.

\bibitem{figueiredo2018high}
W.~Figueiredo, ``A high-speed robotic disassembly system for the recycling and
  reuse of cellphones,'' Ph.D. dissertation, Massachusetts Institute of
  Technology, 2018.

\bibitem{huang2020case}
J.~Huang, D.~T. Pham, Y.~Wang, M.~Qu, C.~Ji, S.~Su, W.~Xu, Q.~Liu, and Z.~Zhou,
  ``A case study in human--robot collaboration in the disassembly of
  press-fitted components,'' \emph{Proceedings of the Institution of Mechanical
  Engineers, Part B: Journal of Engineering Manufacture}, vol. 234, no.~3, pp.
  654--664, 2020.

\bibitem{schmitt2011disassembly}
J.~Schmitt, H.~Haupt, M.~Kurrat, and A.~Raatz, ``Disassembly automation for
  lithium-ion battery systems using a flexible gripper,'' in \emph{2011 15th
  International Conference on Advanced Robotics}.\hskip 1em plus 0.5em minus
  0.4em\relax IEEE, 2011, pp. 291--297.

\bibitem{borras2018kit}
J.~Borras, R.~Heudorfer, S.~Rader, P.~Kaiser, and T.~Asfour, ``The {KIT} swiss
  knife gripper for disassembly tasks: A multi-functional gripper for bimanual
  manipulation with a single arm,'' in \emph{2018 IEEE/RSJ International
  Conference on Intelligent Robots and Systems}.\hskip 1em plus 0.5em minus
  0.4em\relax IEEE, 2018, pp. 4590--4597.

\bibitem{hjorth2022human}
S.~Hjorth and D.~Chrysostomou, ``Human--robot collaboration in industrial
  environments: A literature review on non-destructive disassembly,''
  \emph{Robotics and Computer-Integrated Manufacturing}, vol.~73, p. 102208,
  2022.

\bibitem{huang2021experimental}
J.~Huang, D.~T. Pham, R.~Li, M.~Qu, Y.~Wang, M.~Kerin, S.~Su, C.~Ji,
  O.~Mahomed, R.~Khalil \emph{et~al.}, ``An experimental human-robot
  collaborative disassembly cell,'' \emph{Computers \& Industrial Engineering},
  vol. 155, p. 107189, 2021.

\bibitem{fernandez2006part}
C.~Fernandez, O.~Reinoso, M.~A. Vicente, and R.~Aracil, ``Part grasping for
  automated disassembly,'' \emph{The International Journal of Advanced
  Manufacturing Technology}, vol.~30, pp. 540--553, 2006.

\bibitem{schumacher2014force}
P.~Schumacher and M.~Jouaneh, ``A force sensing tool for disassembly
  operations,'' \emph{Robotics and Computer-Integrated Manufacturing}, vol.~30,
  no.~2, pp. 206--217, 2014.

\bibitem{lee2020making}
M.~A. Lee, Y.~Zhu, P.~Zachares, M.~Tan, K.~Srinivasan, S.~Savarese, L.~Fei-Fei,
  A.~Garg, and J.~Bohg, ``Making sense of vision and touch: Learning multimodal
  representations for contact-rich tasks,'' \emph{IEEE Transactions on
  Robotics}, vol.~36, no.~3, pp. 582--596, 2020.

\bibitem{devo2020towards}
A.~Devo, G.~Mezzetti, G.~Costante, M.~L. Fravolini, and P.~Valigi, ``Towards
  generalization in target-driven visual navigation by using deep reinforcement
  learning,'' \emph{IEEE Transactions on Robotics}, vol.~36, no.~5, pp.
  1546--1561, 2020.

\bibitem{bansal2020combining}
S.~Bansal, V.~Tolani, S.~Gupta, J.~Malik, and C.~Tomlin, ``Combining optimal
  control and learning for visual navigation in novel environments,'' in
  \emph{Conference on Robot Learning}.\hskip 1em plus 0.5em minus 0.4em\relax
  PMLR, 2020, pp. 420--429.

\bibitem{huang2023tracking}
M.~Huang, X.~Li, J.~Hu, H.~Peng, and S.~Lyu, ``Tracking multiple deformable
  objects in egocentric videos,'' in \emph{Proceedings of the IEEE/CVF
  Conference on Computer Vision and Pattern Recognition}, 2023, pp. 1461--1471.

\bibitem{gordon2018re}
D.~Gordon, A.~Farhadi, and D.~Fox, ``$\text{Re}^3$: Real-time recurrent
  regression networks for visual tracking of generic objects,'' \emph{IEEE
  Robotics and Automation Letters}, vol.~3, no.~2, pp. 788--795, 2018.

\bibitem{fan2017parallel}
H.~Fan and H.~Ling, ``Parallel tracking and verifying: A framework for
  real-time and high accuracy visual tracking,'' in \emph{Proceedings of the
  IEEE international conference on computer vision}, 2017, pp. 5486--5494.

\bibitem{kiani2017need}
H.~Kiani~Galoogahi, A.~Fagg, C.~Huang, D.~Ramanan, and S.~Lucey, ``Need for
  speed: A benchmark for higher frame rate object tracking,'' in
  \emph{Proceedings of the IEEE international conference on computer vision},
  2017, pp. 1125--1134.

\bibitem{song2022transformer}
Z.~Song, J.~Yu, Y.-P.~P. Chen, and W.~Yang, ``Transformer tracking with cyclic
  shifting window attention,'' in \emph{Proceedings of the IEEE/CVF conference
  on computer vision and pattern recognition}, 2022, pp. 8791--8800.

\bibitem{richter2021robotic}
F.~Richter, J.~Lu, R.~K. Orosco, and M.~C. Yip, ``Robotic tool tracking under
  partially visible kinematic chain: A unified approach,'' \emph{IEEE
  Transactions on Robotics}, vol.~38, no.~3, pp. 1653--1670, 2021.

\bibitem{marvasti2021deep}
S.~M. Marvasti-Zadeh, L.~Cheng, H.~Ghanei-Yakhdan, and S.~Kasaei, ``Deep
  learning for visual tracking: A comprehensive survey,'' \emph{IEEE
  Transactions on Intelligent Transportation Systems}, vol.~23, no.~5, pp.
  3943--3968, 2021.

\bibitem{horvath2022object}
D.~Horv{\'a}th, G.~Erd{\H{o}}s, Z.~Istenes, T.~Horv{\'a}th, and S.~F{\"o}ldi,
  ``Object detection using sim2real domain randomization for robotic
  applications,'' \emph{IEEE Transactions on Robotics}, vol.~39, no.~2, pp.
  1225--1243, 2022.

\bibitem{sundermeyer2018implicit}
M.~Sundermeyer, Z.-C. Marton, M.~Durner, M.~Brucker, and R.~Triebel,
  ``{Implicit 3D orientation learning for 6D object detection from RGB
  images},'' in \emph{Proceedings of the European Conference on Computer
  Vision}, 2018, pp. 699--715.

\bibitem{wang20206}
C.~Wang, R.~Mart{\'\i}n-Mart{\'\i}n, D.~Xu, J.~Lv, C.~Lu, L.~Fei-Fei,
  S.~Savarese, and Y.~Zhu, ``{6-PACK: Category-level 6D pose tracker with
  anchor-based keypoints},'' in \emph{2020 IEEE International Conference on
  Robotics and Automation}.\hskip 1em plus 0.5em minus 0.4em\relax IEEE, 2020,
  pp. 10\,059--10\,066.

\bibitem{li2018deepim}
Y.~Li, G.~Wang, X.~Ji, Y.~Xiang, and D.~Fox, ``{DeepIM: Deep iterative matching
  for 6D pose estimation},'' in \emph{Proceedings of the European Conference on
  Computer Vision}, 2018, pp. 683--698.

\bibitem{tan2023smoc}
T.~Tan and Q.~Dong, ``{SMOC-Net}: Leveraging camera pose for self-supervised
  monocular object pose estimation,'' in \emph{Proceedings of the IEEE/CVF
  Conference on Computer Vision and Pattern Recognition}, 2023, pp.
  21\,307--21\,316.

\bibitem{al2022robotic}
A.~Al-Shanoon and H.~Lang, ``{Robotic manipulation based on 3-D visual servoing
  and deep neural networks},'' \emph{Robotics and Autonomous Systems}, vol.
  152, p. 104041, 2022.

\bibitem{hay2023noise}
O.~A. Hay, M.~Chehadeh, A.~Ayyad, M.~Wahbah, M.~A. Humais, I.~Boiko,
  L.~Seneviratne, and Y.~Zweiri, ``Noise-tolerant identification and tuning
  approach using deep neural networks for visual servoing applications,''
  \emph{IEEE Transactions on Robotics}, vol.~39, no.~3, pp. 2276--2288, 2023.

\bibitem{bateux2018training}
Q.~Bateux, E.~Marchand, J.~Leitner, F.~Chaumette, and P.~Corke, ``Training deep
  neural networks for visual servoing,'' in \emph{2018 IEEE International
  Conference on Robotics and Automation}.\hskip 1em plus 0.5em minus
  0.4em\relax IEEE, 2018, pp. 3307--3314.

\bibitem{costante2020uncertainty}
G.~Costante and M.~Mancini, ``Uncertainty estimation for data-driven visual
  odometry,'' \emph{IEEE Transactions on Robotics}, vol.~36, no.~6, pp.
  1738--1757, 2020.

\bibitem{fulton2019robotic}
M.~Fulton, J.~Hong, M.~J. Islam, and J.~Sattar, ``Robotic detection of marine
  litter using deep visual detection models,'' in \emph{2019 International
  Conference on Robotics and Automation}.\hskip 1em plus 0.5em minus
  0.4em\relax IEEE, 2019, pp. 5752--5758.

\bibitem{zocco2023towards}
F.~Zocco, T.-C. Lin, C.-I. Huang, H.-C. Wang, M.~O. Khyam, and M.~Van,
  ``Towards more efficient {EfficientDets} and real-time marine debris
  detection,'' \emph{IEEE Robotics and Automation Letters}, vol.~8, no.~4, pp.
  2134--2141, 2023.

\bibitem{hong2020trashcan}
J.~Hong, M.~Fulton, and J.~Sattar, ``{TrashCan}: A semantically-segmented
  dataset towards visual detection of marine debris,'' \emph{arXiv preprint
  arXiv:2007.08097}, 2020.

\bibitem{bashkirova2022zerowaste}
D.~Bashkirova, M.~Abdelfattah, Z.~Zhu, J.~Akl, F.~Alladkani, P.~Hu,
  V.~Ablavsky, B.~Calli, S.~A. Bargal, and K.~Saenko, ``{ZeroWaste} dataset:
  Towards deformable object segmentation in cluttered scenes,'' in
  \emph{Proceedings of the IEEE/CVF Conference on Computer Vision and Pattern
  Recognition}, 2022, pp. 21\,147--21\,157.

\bibitem{bai2018deep}
J.~Bai, S.~Lian, Z.~Liu, K.~Wang, and D.~Liu, ``Deep learning based robot for
  automatically picking up garbage on the grass,'' \emph{IEEE Transactions on
  Consumer Electronics}, vol.~64, no.~3, pp. 382--389, 2018.

\bibitem{wang2020vision}
Z.~Wang, H.~Li, and X.~Yang, ``Vision-based robotic system for on-site
  construction and demolition waste sorting and recycling,'' \emph{Journal of
  Building Engineering}, vol.~32, p. 101769, 2020.

\bibitem{kiyokawa2021robotic}
T.~Kiyokawa, H.~Katayama, Y.~Tatsuta, J.~Takamatsu, and T.~Ogasawara, ``Robotic
  waste sorter with agile manipulation and quickly trainable detector,''
  \emph{IEEE Access}, vol.~9, pp. 124\,616--124\,631, 2021.

\bibitem{jahanian2019see}
A.~Jahanian, Q.~H. Le, K.~Youcef-Toumi, and D.~Tsetserukou, ``See the e-waste!
  {Training} visual intelligence to see dense circuit boards for recycling,''
  in \emph{Proceedings of the IEEE/CVF Conference on Computer Vision and
  Pattern Recognition Workshops}, 2019, pp. 1--8.

\bibitem{brogan2021deep}
D.~P. Brogan, N.~M. DiFilippo, and M.~K. Jouaneh, ``Deep learning computer
  vision for robotic disassembly and servicing applications,'' \emph{Array},
  vol.~12, p. 100094, 2021.

\bibitem{mangold2022vision}
S.~Mangold, C.~Steiner, M.~Friedmann, and J.~Fleischer, ``Vision-based screw
  head detection for automated disassembly for remanufacturing,''
  \emph{Procedia CIRP}, vol. 105, pp. 1--6, 2022.

\bibitem{foo2021screw}
G.~Foo, S.~Kara, and M.~Pagnucco, ``Screw detection for disassembly of
  electronic waste using reasoning and re-training of a deep learning model,''
  \emph{Procedia CIRP}, vol.~98, pp. 666--671, 2021.

\bibitem{zocco2023visual}
F.~Zocco and S.~Rahimifard, ``Visual material characteristics learning for
  circular healthcare,'' \emph{arXiv preprint arXiv:2309.04763}, 2023.

\bibitem{zocco2024towards}
F.~Zocco, D.~Sleath, and S.~Rahimifard, ``Towards a thermodynamical
  deep-learning-vision-based flexible robotic cell for circular healthcare,''
  \emph{arXiv preprint arXiv:2402.05551}, 2024.

\bibitem{zocco2022material}
F.~Zocco, S.~McLoone, and B.~Smyth, ``Material measurement units for a circular
  economy: Foundations through a review,'' \emph{Sustainable Production and
  Consumption}, vol.~32, pp. 833--850, 2022.

\bibitem{bell2015material}
S.~Bell, P.~Upchurch, N.~Snavely, and K.~Bala, ``Material recognition in the
  wild with the materials in context database,'' in \emph{Proceedings of the
  IEEE Conference on Computer Vision and Pattern Recognition}, 2015, pp.
  3479--3487.

\bibitem{schwartz2019recognizing}
G.~Schwartz and K.~Nishino, ``Recognizing material properties from images,''
  \emph{IEEE Transactions on Pattern Analysis and Machine Intelligence},
  vol.~42, no.~8, pp. 1981--1995, 2019.

\bibitem{lagunas2019similarity}
M.~Lagunas, S.~Malpica, A.~Serrano, E.~Garces, D.~Gutierrez, and B.~Masia, ``A
  similarity measure for material appearance,'' \emph{ACM Transactions on
  Graphics}, vol.~38, no.~4, pp. 1--12, 2019.

\bibitem{zocco2024synchronized}
F.~Zocco, D.~Lake, and S.~Rahimifard, ``Synchronized object detection for
  autonomous sorting, mapping, and quantification of medical materials,''
  \emph{arXiv preprint arXiv:2405.06821}, 2024.

\bibitem{standley2017image2mass}
T.~Standley, O.~Sener, D.~Chen, and S.~Savarese, ``image2mass: Estimating the
  mass of an object from its image,'' in \emph{Conference on Robot
  Learning}.\hskip 1em plus 0.5em minus 0.4em\relax PMLR, 2017, pp. 324--333.

\bibitem{andrade2023improving}
J.~M.~L. Andrade and P.~Moreno, ``Improving the estimation of object mass from
  images,'' in \emph{2023 IEEE International Conference on Autonomous Robot
  Systems and Competitions}.\hskip 1em plus 0.5em minus 0.4em\relax IEEE, 2023,
  pp. 199--206.

\bibitem{patel2022adding}
D.~Patel, A.~Nath, and R.~Niyogi, ``Adding material embedding to the image2mass
  problem,'' in \emph{International Conference on Computational Science and Its
  Applications}.\hskip 1em plus 0.5em minus 0.4em\relax Springer, 2022, pp.
  77--90.

\bibitem{diaz2022simultaneous}
D.~J. D{\'\i}az-Romero, S.~Van~den Eynde, W.~Sterkens, B.~Engelen, I.~Zaplana,
  W.~Dewulf, T.~Goedem{\'e}, and J.~Peeters, ``Simultaneous mass estimation and
  class classification of scrap metals using deep learning,'' \emph{Resources,
  Conservation and Recycling}, vol. 181, p. 106272, 2022.

\bibitem{moran2010fundamentals}
M.~J. Moran, H.~N. Shapiro, D.~D. Boettner, and M.~B. Bailey,
  \emph{Fundamentals of Engineering Thermodynamics}.\hskip 1em plus 0.5em minus
  0.4em\relax John Wiley \& Sons, 2010.

\bibitem{liu2021garbage}
J.~Liu, P.~Balatti, K.~Ellis, D.~Hadjivelichkov, D.~Stoyanov, A.~Ajoudani, and
  D.~Kanoulas, ``Garbage collection and sorting with a mobile manipulator using
  deep learning and whole-body control,'' in \emph{2020 IEEE-RAS 20th
  International Conference on Humanoid Robots (Humanoids)}.\hskip 1em plus
  0.5em minus 0.4em\relax IEEE, 2021, pp. 408--414.

\bibitem{gioioso2014flying}
G.~Gioioso, A.~Franchi, G.~Salvietti, S.~Scheggi, and D.~Prattichizzo, ``The
  flying hand: A formation of {UAVs} for cooperative aerial
  tele-manipulation,'' in \emph{2014 IEEE International Conference on Robotics
  and Automation}.\hskip 1em plus 0.5em minus 0.4em\relax IEEE, 2014, pp.
  4335--4341.

\bibitem{six2017kinematics}
D.~Six, S.~Briot, A.~Chriette, and P.~Martinet, ``The kinematics, dynamics and
  control of a flying parallel robot with three quadrotors,'' \emph{IEEE
  Robotics and Automation Letters}, vol.~3, no.~1, pp. 559--566, 2017.

\bibitem{lim2021microplastics}
X.~Lim \emph{et~al.}, ``Microplastics are everywhere—but are they harmful,''
  \emph{Nature}, vol. 593, no. 7857, pp. 22--25, 2021.

\bibitem{sagatun1991lagrangian}
S.~I. Sagatun and T.~I. Fossen, ``Lagrangian formulation of underwater
  vehicles' dynamics,'' in \emph{Conference Proceedings 1991 IEEE International
  Conference on Systems, Man, and Cybernetics}.\hskip 1em plus 0.5em minus
  0.4em\relax IEEE, 1991, pp. 1029--1034.

\bibitem{bandh2023biofuels}
S.~A. Bandh and F.~A. Malla, \emph{Biofuels in Circular Economy}.\hskip 1em
  plus 0.5em minus 0.4em\relax Springer Nature, 2023.

\bibitem{bernard2001dynamical}
O.~Bernard, Z.~Hadj-Sadok, D.~Dochain, A.~Genovesi, and J.-P. Steyer,
  ``Dynamical model development and parameter identification for an anaerobic
  wastewater treatment process,'' \emph{Biotechnology and Bioengineering},
  vol.~75, no.~4, pp. 424--438, 2001.

\bibitem{campos2019hybrid}
A.~Campos-Rodr{\'\i}guez, J.~Garc{\'\i}a-Sandoval, V.~Gonz{\'a}lez-{\'A}lvarez,
  and A.~Gonz{\'a}lez-{\'A}lvarez, ``Hybrid cascade control for a class of
  nonlinear dynamical systems,'' \emph{Journal of Process Control}, vol.~76,
  pp. 141--154, 2019.

\bibitem{mcdonald1989thermal}
A.~T. McDonald, S.~H. Friskney, and D.~J. Ulrich, ``Thermal model of the
  dishwasher heater in air,'' \emph{IEEE Transactions on Industry
  Applications}, vol.~25, no.~6, pp. 1176--1180, 1989.

\bibitem{magnanelli2020dynamic}
E.~Magnanelli, O.~L. Tran{\aa}s, P.~Carlsson, J.~Mosby, and M.~Becidan,
  ``Dynamic modeling of municipal solid waste incineration,'' \emph{Energy},
  vol. 209, p. 118426, 2020.

\bibitem{kumar2022design}
A.~Kumar and S.~K. Verma, ``Design and development of e-smart robotics-based
  underground solid waste storage and transportation system,'' \emph{Journal of
  Cleaner Production}, vol. 343, p. 130987, 2022.

\bibitem{sarc2019digitalisation}
R.~Sarc, A.~Curtis, L.~Kandlbauer, K.~Khodier, K.~E. Lorber, and R.~Pomberger,
  ``Digitalisation and intelligent robotics in value chain of circular economy
  oriented waste management--{A} review,'' \emph{Waste Management}, vol.~95,
  pp. 476--492, 2019.

\bibitem{lin2024development}
Y.-H. Lin, W.-L. Mao, and H.~I.~K. Fathurrahman, ``Development of intelligent
  municipal solid waste sorter for recyclables,'' \emph{Waste Management}, vol.
  174, pp. 597--604, 2024.

\bibitem{gundupalli2017review}
S.~P. Gundupalli, S.~Hait, and A.~Thakur, ``A review on automated sorting of
  source-separated municipal solid waste for recycling,'' \emph{Waste
  management}, vol.~60, pp. 56--74, 2017.

\bibitem{aschenbrenner2023robot}
D.~Aschenbrenner, C.~Colloseus, R.~Khoury, and N.~Fangerow, ``Robot-assisted
  automated sorting techniques for plastic recycling,'' \emph{Procedia CIRP},
  vol. 120, pp. 1232--1237, 2023.

\bibitem{chen2022robot}
X.~Chen, H.~Huang, Y.~Liu, J.~Li, and M.~Liu, ``Robot for automatic waste
  sorting on construction sites,'' \emph{Automation in Construction}, vol. 141,
  p. 104387, 2022.

\bibitem{BritishPF}
Payal\hspace{3pt}Baheti\hspace{3pt}at\hspace{3pt}British\hspace{3pt}Plastics\hspace{3pt}Federation,
  ``How is plastic made? {A} simple step-by-step explanation,'' 2024, \,
  available at:
  \url{https://www.bpf.co.uk/plastipedia/how-is-plastic-made.aspx#:~:text=In%20the%20refining%20process%2C%20crude,means%2C%20such%20as%20using%20gas.};
  last access: 19 September 2024.

\bibitem{BioRobWeb1}
Circular\hspace{3pt}Bio-based\hspace{3pt}Europe, ``{ReBioCycle: A new European
  blueprint for circular bioplastics upcycling solutions},'' 2024, \, available
  at: \url{https://www.cbe.europa.eu/projects/rebiocycle}; last access: 19
  September 2024.

\bibitem{BioRobWeb2}
Iren\hspace{3pt}Group, ``{EU-DREAM e ReBioCycle, i progetti in partnership con
  Iren finanziati dalla Commissione europea},'' 2024, \, available at:
  \url{https://www.gruppoiren.it/it/everyday/sfide-di-innovazione/2024/progetti-partnership-iren-finanziamenti-commissione-europea.html};
  last access: 19 September 2024.

\bibitem{Goodfellow-et-al-2016}
I.~Goodfellow, Y.~Bengio, and A.~Courville, \emph{Deep Learning}.\hskip 1em
  plus 0.5em minus 0.4em\relax MIT Press, 2016,
  \url{http://www.deeplearningbook.org}.

\bibitem{lillicrap2015continuous}
T.~Lillicrap, ``Continuous control with deep reinforcement learning,''
  \emph{arXiv preprint arXiv:1509.02971}, 2015.

\bibitem{A2C}
Stable\hspace{3pt}Baselines3, 2024, \, webpage:
  \url{https://stable-baselines3.readthedocs.io/en/master/modules/a2c.html};
  last access: 24 September 2024.

\bibitem{schulman2017proximal}
J.~Schulman, F.~Wolski, P.~Dhariwal, A.~Radford, and O.~Klimov, ``Proximal
  policy optimization algorithms,'' \emph{arXiv preprint arXiv:1707.06347},
  2017.

\bibitem{haarnoja2018soft}
T.~Haarnoja, A.~Zhou, P.~Abbeel, and S.~Levine, ``Soft actor-critic:
  {Off-policy maximum entropy deep reinforcement learning with a stochastic
  actor},'' in \emph{International Conference on Machine Learning}.\hskip 1em
  plus 0.5em minus 0.4em\relax PMLR, 2018, pp. 1861--1870.

\bibitem{MuJoCo-Robots}
Gymnasium\hspace{3pt}documentation, 2023, \, webpage:
  \url{https://gymnasium.farama.org/environments/mujoco/reacher/}; last access:
  24 September 2024.

\end{thebibliography}

\begin{IEEEbiography}[{\includegraphics[width=1in,height=1.25in,clip,keepaspectratio]{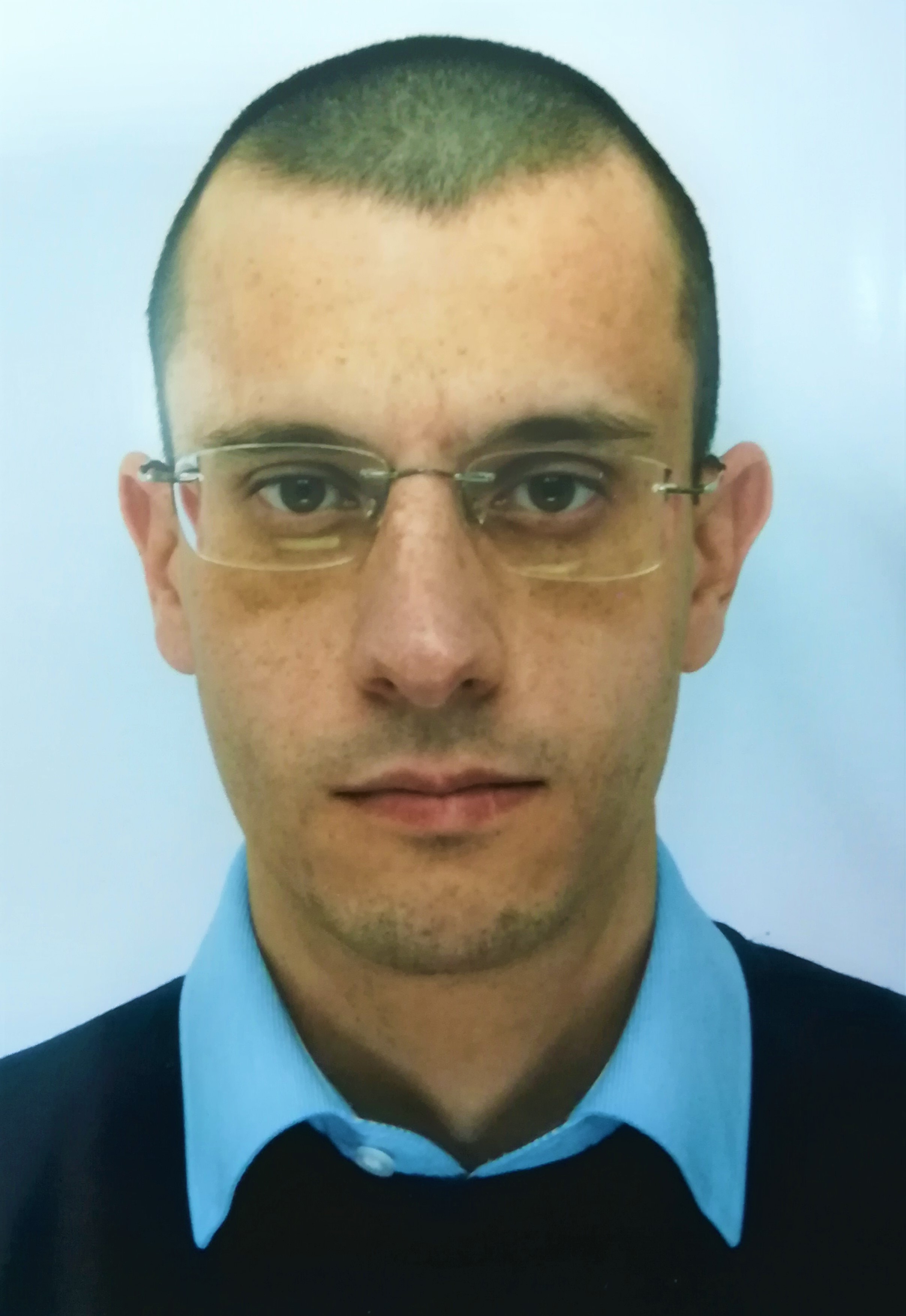}}]{Federico Zocco} received the B.S. in Mechanical Engineering and the M.S. in Robotics and Automation Engineering from University of Pisa in 2013 and 2016, respectively, and the Ph.D. in Applied Machine Learning from Queen’s University Belfast in 2021. He is currently a postdoctoral researcher with the School of Mechanical, Electrical and Manufacturing Engineering and with the Centre for Sustainable Manufacturing and Recycling Technologies both at Loughborough University.

Dr. Zocco’s research merges computer vision, compartmental dynamical thermodynamics, automatic control theory, and network science with the holistic perspective of industrial ecology to define the theoretical foundations of circular flow designs of materials. The main area of application is the circularity of critical materials such as the atmospheric carbon dioxide and the rare-earth metals.
 
\end{IEEEbiography}

\begin{IEEEbiography}[{\includegraphics[width=1in,height=1.25in,clip,keepaspectratio]{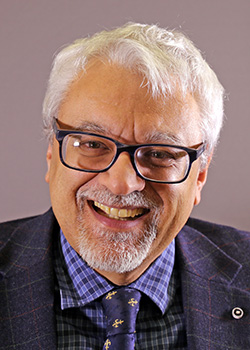}}]{Wassim M. Haddad} received the B.S., M.S., and Ph.D. degrees in mechanical engineering from Florida Tech in 1983, 1984, and 1987. From 1987 to 1994 he served as a consultant for the Structural Controls Group of the Government Aerospace Systems Division, Harris Corporation, Melbourne, FL. In 1988 he joined the faculty of the Mechanical and Aerospace Engineering Department at Florida Tech, where he founded and developed the Systems and Control Option within the graduate program. Since 1994 he has been with the School of Aerospace Engineering at Georgia Tech, where he holds the rank of Professor, the David Lewis Chair in Dynamical Systems and Control, and Chair of the Flight Mechanics and Control Discipline. He also holds a joint Professor appointment with the School of Electrical and Computer Engineering at Georgia Tech. He is the Co-Founder, Chairman of the Board, and Chief Scientific Advisor of Autonomous Healthcare, Inc. 

Dr. Haddad has made numerous contributions to the development of nonlinear control theory and its application to aerospace, electrical, and biomedical engineering. His transdisciplinary research in dynamical systems and control is documented in over 700 archival journal and conference publications, and 8 books in the areas of science, mathematics, medicine, and engineering. Dr. Haddad is an NSF Presidential Faculty Fellow; a member of the Academy of Nonlinear Sciences; an IEEE Fellow; an AAIA Fellow; and the recipient of the AIAA Pendray Aerospace Literature Award. 
\end{IEEEbiography}

\begin{IEEEbiography}
[{\includegraphics[width=1in,height=1.25in,clip,keepaspectratio]{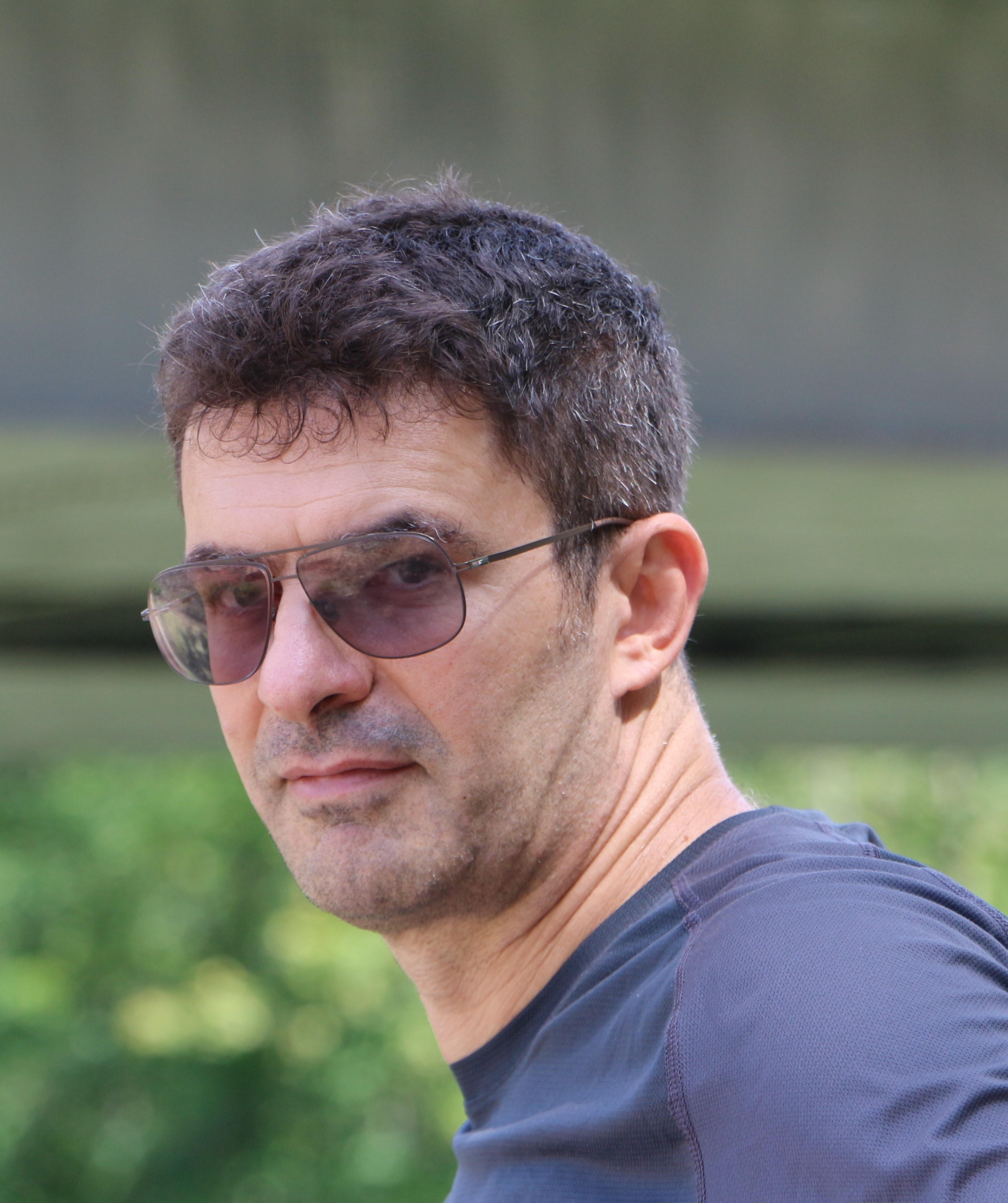}}]{Andrea Corti} is Associate Professor of Energy and Environmental Systems at the University of Siena, Italy. He earned the Ph.D. degree in energy systems on low CO$_2$ energy conversion from the University of Florence in 1998. He was a contract researcher at the University of Naples and University of Florence on environmental technologies applied to energy conversion systems.

His main research interests include applied thermodynamics, simulation of Environmental Impact from energy systems, renewables technologies, biological process for biogas production energy, and circular economy technologies.
He is author of several publications in journals, international conferences, and book chapters.
\end{IEEEbiography}

\begin{IEEEbiography}
[{\includegraphics[width=1in,height=1.25in,clip,keepaspectratio]{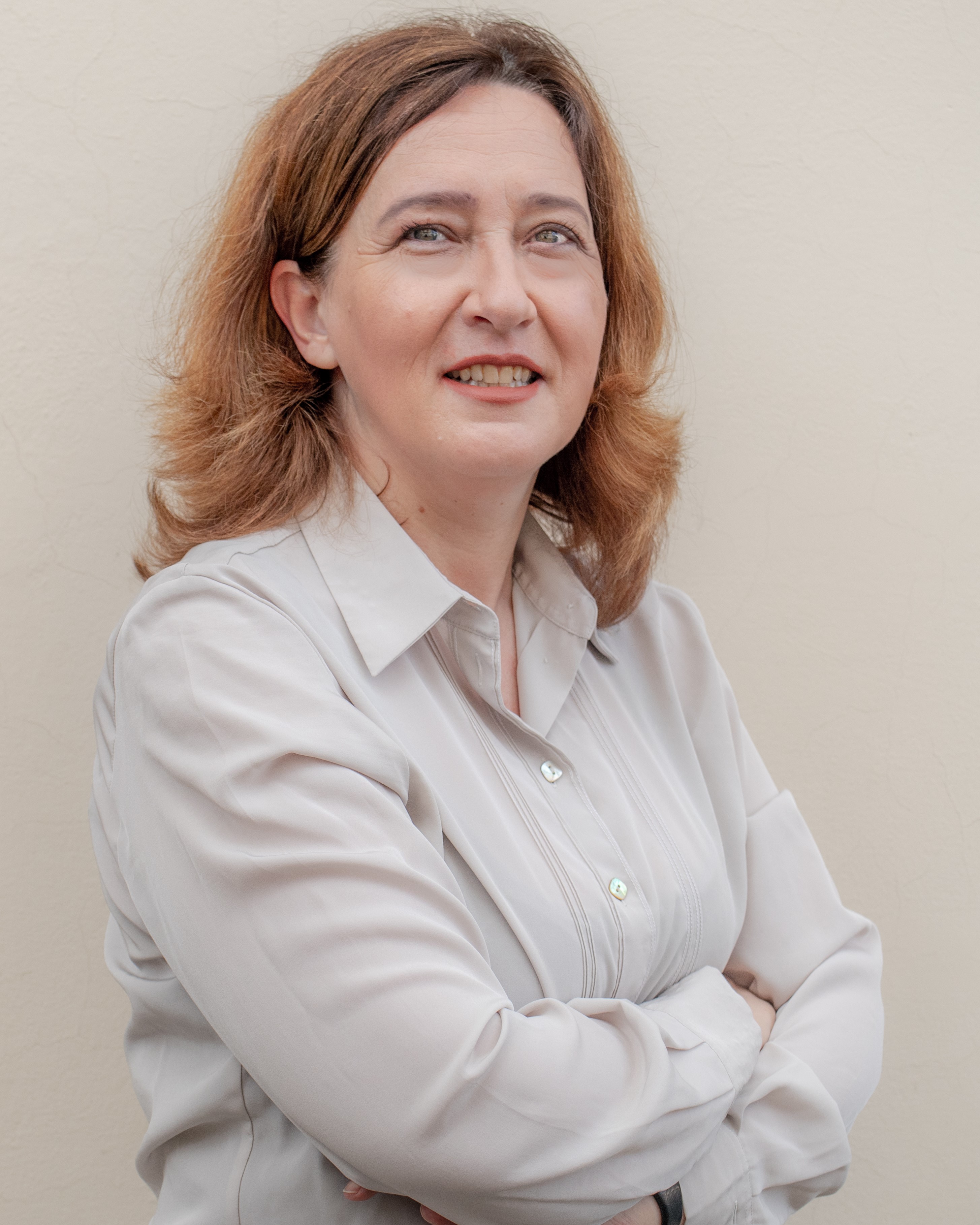}}]{Monica Malvezzi} is Associate Professor of mechanics and mechanism theory at the University of Siena, Italy. She earned her Ph.D. degree in applied mechanics from the University of Bologna in 2003. She has also been Assistant Professor at the University of Siena from 2008 to 2018, Researcher at the University of Florence from 2002 to 2008 and Visiting Scientist at the Department of Advanced Robotics, Istituto Italiano di Tecnologia, Genova, from 2015 to 2019. 
She was PI for the UNISI unit in the H2020 project ``INBOTS - Inclusive Robotics for a better Society'' and in the ERASMUS+ project BEREADY.

Her main research interests include the control of mechanical and mechatronic systems, robotics, haptics, multibody dynamics, grasping, and dexterous manipulation. She is the author of several publications in journals, international conferences, and book chapters and serves as member of the editorial/organizing board of international conferences and journals. 
\end{IEEEbiography}

\EOD
\end{document}